%% file: iclr2027_conference.tex
\documentclass{article} 
\usepackage{iclr2027_conference,times}

\input{math_commands.tex}

\usepackage{hyperref}
\usepackage{url}

\usepackage{booktabs}
\usepackage{tabularx}
\usepackage{array}
\usepackage{makecell}
\usepackage{pifont}
\usepackage{float}
\usepackage{multirow}

\usepackage{xcolor}
\usepackage{colortbl}
\usepackage{graphicx}
\usepackage{cuted}

\usepackage{caption}

\definecolor{scopeblue}{RGB}{226,238,249}
\definecolor{sectionblue}{RGB}{239,246,252}

\definecolor{topogreen}{RGB}{228,242,222}
\definecolor{sectiongreen}{RGB}{240,248,237}

\definecolor{oursyellow}{RGB}{255,244,190}

\definecolor{primarygreen}{RGB}{22,122,72}
\definecolor{partialorange}{RGB}{204,119,34}

\providecommand{\Primary}{%
    \textcolor{primarygreen}{\ding{51}}%
}

\providecommand{\Partial}{%
    \textcolor{partialorange}{$\triangle$}%
}

\providecommand{\NoMark}{%
    \textcolor{gray!45}{--}%
}

\author{
Songlin Duan\textsuperscript{1,*},
Shuo Yang\textsuperscript{1,*},
Yujia Zheng\textsuperscript{2,\dag},
Tobin Filleter\textsuperscript{1,\dag} \\
\textsuperscript{1}University of Toronto, Canada 
\textsuperscript{2}University of Illinois Urbana-Champaign, USA
}

\iclrfinalcopy 
\begin{document}

\title{Learning Topology-Aware Representations for Mechanics-Relevant TPMS Metamaterials}

\begingroup
\renewcommand{\thefootnote}{}
\footnotetext{\textsuperscript{*}Both authors contributed equally to this work.}
\footnotetext{\textsuperscript{\dag}Corresponding authors.}
\endgroup

\makeatletter
\renewcommand{\@oddhead}{}
\renewcommand{\@evenhead}{}
\makeatother

\maketitle

\begin{abstract}
Learning informative representations of mechanical Triply Periodic Minimal Surface (TPMS) metamaterials is important for mechanics analysis and structural design, yet existing methods have rarely explicitly supervised learned representations with information about global structural connectivity that is relevant to mechanical response. We propose a topology-aware representation learning framework that integrates persistent homology into a DMAE backbone to explicitly encode global structural connectivity. A persistence-guided masking strategy prioritizes patches associated with persistent topological features, while an auxiliary persistent-homology decoder encourages the latent representation to retain information about topological feature counts across connected components, tunnels, and voids. We evaluate the learned representations on two complementary TPMS datasets using family classification, mechanical-property classification, unsupervised clustering, and regression. Across the two datasets, our method demonstrates strong performance in both structural family and mechanical-property classification, improving family classification by up to $3.9\%$ while achieving the best linear-probe performance. More notably, unsupervised property-clustering performance improves by up to $26.7\%$ and $51.5\%$ across the two datasets, while strong regression performance is retained with $R^2=0.993$. These results establish persistent topology as an effective inductive bias for representation learning of complex three-dimensional mechanical metamaterials.

\end{abstract}

\begin{figure}[h]
    \centering
    \includegraphics[width=0.72\linewidth]{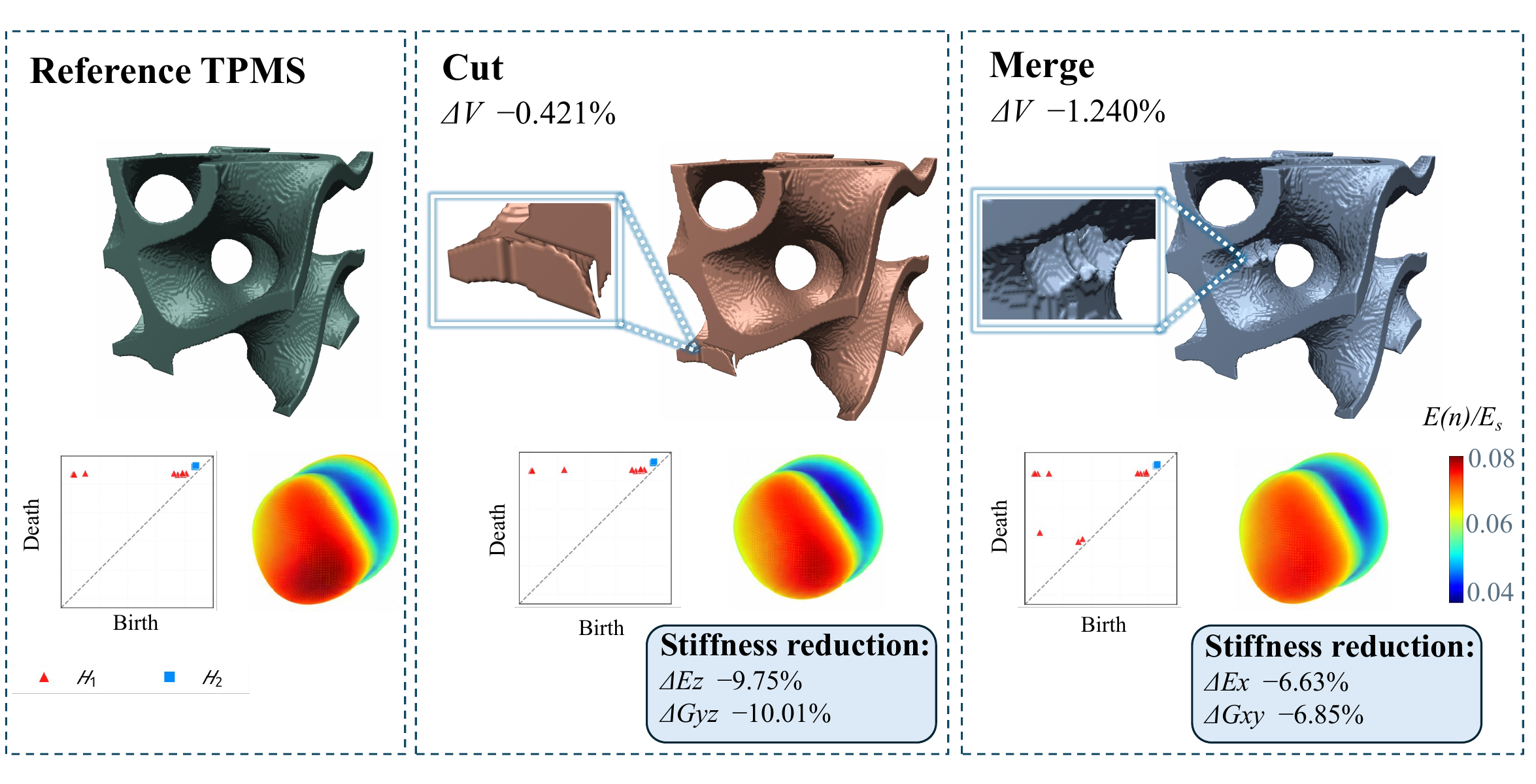}
    \caption{
    \textbf{Localized topological changes can induce disproportionate changes in the mechanical response of TPMS structures.}
    Starting from the reference TPMS (left), the \textit{Cut} (middle) introduces a physical crack at an existing structural connection, whereas the \textit{Merge} (right) joins two separated tunnels into a single connected passage. Despite only small changes in solid volume fraction ($\Delta V=-0.421\%$ and $-1.240\%$, respectively), both modifications produce clear changes in the persistence diagrams (PDs, summaries of topological features) and directional elastic responses. The \textit{Cut} yields maximum reductions of $9.75\%$ in $E_z$ and $10.01\%$ in $G_{yz}$, whereas the \textit{Merge} reduces $E_x$ by $6.63\%$ and $G_{xy}$ by $6.85\%$. Additional topological perturbations, full directional stiffness changes, and detailed analysis are provided in Appendix~\ref{sec:topo_sensitivity} and Fig.~\ref{fig:topology_mechanics}.
    }
    \label{fig:cut_merge}
\end{figure}

\section{Introduction}

Architected meta-materials enable tailored and unconventional physical responses through structural architecture rather than material composition alone, creating opportunities in lightweight structures~\citep{portela2021supersonic}, energy absorption~\citep{yang2022damage,fang2025large,liu2024harnessing}, thermal management~\citep{li2021transforming}, wave control~\citep{shaltout2019spatiotemporal}, sensing~\citep{berman2024additively}, and other multifunctional applications~\citep{hou2026generative}. Among them, TPMS metamaterials have emerged as a versatile class of architected materials because their smooth, continuous and highly tunable periodic architectures combine interconnected load-bearing and transport pathways with high surface-area-to-volume ratios~\citep{wang2025harnessing}, making them attractive for a wide range of applications, particularly in mechanical metamaterial systems~\citep{zhang2026multiscale,li2025additively,li2018nanophotonic}, as illustrated in Appendix~\ref{sec:tpms_shell_background}.

As increasingly large libraries of mechanical metamaterials are explored computationally~\citep{maurizi2025graphmetamat,beatson2020composable,wan2022visual}, learning compact and informative representations of these complex structures has become increasingly important, and such representations should capture not only geometric similarity but also structural characteristics relevant to mechanical response. Existing studies have learned compact representations for several classes of metamaterials, including periodic microstructures with latent spaces organized by mechanical behavior~\citep{wang2020deep}, nature-inspired architected structures optimized with respect to mechanical properties~\citep{shen2022nature}, and truss-based metamaterials for generation and mechanical-property prediction~\citep{zhan2025unimate,zheng2023unifying}. In contrast, TPMS-specific studies have primarily focused on multimodal generation~\citep{makatura2025metagen}, surrogate-based simulation~\citep{chen2024parameterized}, and multi-objective mechanical optimization~\citep{yan2026multiperformance}, while mechanics-relevant TPMS representation learning has not been explicitly studied as a primary objective.

To better learn mechanics-relevant representations of TPMS structures, it is important to capture not only local geometric characteristics, such as wall thickness, curvature, loops, voids, and surface orientation~\citep{qiu2024experimental}, but also global connectivity through how local features are organized across the structure~\citep{chen2019stiff,gibson1999cellular,deshpande2001foam}. This is consistent with Maxwell-type rigidity arguments, which link structural connectivity and constraint density to accessible deformation modes and mechanical stability~\citep{pellegrino1986matrix}.

Topological data analysis (TDA) provides a natural way to characterize global structural organization, while persistent homology (PH) captures such topology across multiple scales through persistence diagrams (PDs)~\citep{ghrist2008barcodes,edelsbrunner2010computational,edelsbrunner2002topological,zomorodian2005computing}. This characterization is suitable for TPMS geometries, in which small geometric variations can alter global connectivity and create or eliminate topological features, potentially resulting in different mechanical responses~\citep{wong2025scalabletopo,edelsbrunner2002topological}, as illustrated in Figure.~\ref{fig:cut_merge}. Prior topology-aware methods, as summarized in Table~\ref{tab:literature}, have largely incorporated PH to encode structural connectivity in 2D~\citep{hu2023probabilistic,wang2020topogan} or in generic geometric settings~\citep{hofer2019connectivity,moor2020topological,chen2023topologicalpooling,tola2025toper}. In materials and mechanics, PH has been used to characterize crystalline compounds~\citep{jiang2021topological}, structural materials~\citep{tiong2023failure}, granular media~\citep{mei2023modeling}, and amorphous solids~\citep{minamitani2025persistent}. In architected materials, prior studies have mainly used PH as a geometric descriptor or generative condition~\citep{yan2023thickness,gao2025persistent,du2026topogenmeta}, rather than directly supervising the learned latent representation to encode persistent-topological information. Whether such explicit topology preservation can improve mechanics-relevant representations of complex 3D TPMS structures therefore remains largely unexplored.

\begin{table}[t]

\caption{Comparison of representative work across metamaterial and TPMS modeling, mechanics-aware methods, representation learning model, and topology-aware structural learning.}
\label{tab:literature}

\vspace{\baselineskip}

\begin{center}

\scriptsize
\setlength{\tabcolsep}{2.2pt}
\renewcommand{\arraystretch}{1.05}

\begin{tabularx}{\textwidth}{
    >{\raggedright\arraybackslash}X
    *{8}{c}
}

\toprule

\textbf{Representative work}
&
\multicolumn{5}{c}{
    \cellcolor{scopeblue}
    \textbf{\textsc{Scope / focus}}
}
&
\multicolumn{3}{c}{
    \cellcolor{topogreen}
    \textbf{\textsc{Topology-aware}}
}
\\

\cmidrule(lr){2-6}
\cmidrule(lr){7-9}

&
\cellcolor{scopeblue}
\makecell{\textbf{Meta.}}
&
\cellcolor{scopeblue}
\makecell{\textbf{TPMS}}
&
\cellcolor{scopeblue}
\makecell{\textbf{3D}\\\textbf{struct.}}
&
\cellcolor{scopeblue}
\makecell{\textbf{Mech.}\\\textbf{rel.}}
&
\cellcolor{scopeblue}
\makecell{\textbf{Rep.}\\\textbf{learn.}}
&
\cellcolor{topogreen}
\makecell{\textbf{TDA}\\\textbf{/ PH}}
&
\cellcolor{topogreen}
\makecell{\textbf{Topo}\\\textbf{head}}
&
\cellcolor{topogreen}
\makecell{\textbf{Topo}\\\textbf{mask}}
\\

\midrule

\rowcolor{sectionblue}
\multicolumn{9}{l}{
\textbf{\textit{General metamaterial modeling}}
}
\\

GraphMetaMat~\citep{maurizi2025graphmetamat}
&
\Primary & \NoMark & \Primary & \Primary & \NoMark
& \NoMark & \NoMark & \NoMark
\\

Composable Energy Surrogates~\citep{beatson2020composable}
&
\Primary & \NoMark & \Primary & \Primary & \NoMark
& \NoMark & \NoMark & \NoMark
\\

3D Metamaterial Force Control~\citep{wan2022visual}
&
\Primary & \NoMark & \Primary & \Primary & \NoMark
& \NoMark & \NoMark & \NoMark
\\

\addlinespace[1.5pt]

\rowcolor{sectionblue}
\multicolumn{9}{l}{
\textbf{\textit{Metamaterial representation \& inverse design}}
}
\\

Neural-Adjoint~\citep{ren2020benchmarking}
&
\Partial & \NoMark & \NoMark & \Primary & \Partial
& \NoMark & \NoMark & \NoMark
\\

Video Diffusion Metamaterials~\citep{bastek2023inverse}
&
\Primary & \NoMark & \NoMark & \Primary & \Partial
& \NoMark & \NoMark & \NoMark
\\

Deep Generative Metamaterial Design~\citep{wang2020deep}
&
\Primary & \NoMark & \NoMark & \Primary & \Primary
& \NoMark & \NoMark & \NoMark
\\

Nature-Inspired Unsupervised Design~\citep{shen2022nature}
&
\Primary & \NoMark & \Partial & \Primary & \Primary
& \NoMark & \NoMark & \NoMark
\\

UniMate~\citep{zhan2025unimate}
&
\Primary & \NoMark & \Primary & \Primary & \Primary
& \NoMark & \NoMark & \NoMark
\\

Generative Truss Metamaterials~\citep{zheng2023unifying}
&
\Primary & \NoMark & \Primary & \Primary & \Primary
& \NoMark & \NoMark & \NoMark
\\

MetamatBench~\citep{chen2025metamatbench}
&
\Primary & \NoMark & \Primary & \Primary & \Partial
& \NoMark & \NoMark & \NoMark
\\

Property-Informed Diffusion~\citep{dai2026property}
&
\Primary & \NoMark & \Primary & \Primary & \Partial
& \NoMark & \NoMark & \NoMark
\\

\addlinespace[1.5pt]

\rowcolor{sectionblue}
\multicolumn{9}{l}{
\textbf{\textit{TPMS-specific modeling \& optimization}}
}
\\

MetaGen~\citep{makatura2025metagen}
&
\Primary & \Partial & \Primary & \Primary & \Partial
& \NoMark & \NoMark & \NoMark
\\

PNN-TPMS Multiscale FEM~\citep{chen2024parameterized}
&
\Primary & \Primary & \Primary & \Primary & \NoMark
& \NoMark & \NoMark & \NoMark
\\

ML-Optimized TPMS Lattices~\citep{yan2026multiperformance}
&
\Primary & \Primary & \Primary & \Primary & \NoMark
& \NoMark & \NoMark & \NoMark
\\

\rowcolor{sectionblue!45!sectiongreen}

PH-Based Thickness Design~\citep{yan2023thickness}
&
\Partial & \Partial & \Primary & \NoMark & \NoMark
& \Primary & \NoMark & \NoMark
\\

\rowcolor{sectionblue!45!sectiongreen}

PH-Driven TPMS Optimization~\citep{gao2025persistent}
&
\Primary & \Primary & \Primary & \Primary & \NoMark
& \Primary & \NoMark & \NoMark
\\

\addlinespace[1.5pt]

\rowcolor{sectiongreen}
\multicolumn{9}{l}{
\textbf{\textit{Topology-aware modeling \&  learning models}}
}
\\

Topo-GenMeta~\citep{du2026topogenmeta}
&
\Primary & \NoMark & \NoMark & \Primary & \Partial
& \Primary & \NoMark & \NoMark
\\

Topological Crystal Representations~\citep{jiang2021topological}
&
\NoMark & \NoMark & \Primary & \Primary & \Primary
& \Primary & \NoMark & \NoMark
\\

Void-Topology Failure Prediction~\citep{tiong2023failure}
&
\NoMark & \NoMark & \Primary & \Primary & \NoMark
& \Primary & \NoMark & \NoMark
\\

PH for Granular Transition~\citep{mei2023modeling}
&
\NoMark & \NoMark & \Primary & \Primary & \NoMark
& \Primary & \NoMark & \NoMark
\\

PH for Amorphous Solids~\citep{minamitani2025persistent}
&
\NoMark & \NoMark & \Primary & \Primary & \NoMark
& \Primary & \NoMark & \NoMark
\\

Conformable Convolution~\citep{yeganeh2026conformable}
&
\NoMark & \NoMark & \Partial & \NoMark & \NoMark
& \Primary & \NoMark & \NoMark
\\

Multiparameter PH Representations~\citep{loiseaux2023framework}
&
\NoMark & \NoMark & \Partial & \NoMark & \Primary
& \Primary & \NoMark & \NoMark
\\

Point-Level TopoRL~\citep{grande2025point}
&
\NoMark & \NoMark & \Primary & \NoMark & \NoMark
& \Primary & \NoMark & \NoMark
\\

Probabilistic Topological Representations~\citep{hu2023probabilistic}
&
\NoMark & \NoMark & \NoMark & \NoMark & \Primary
& \Primary & \NoMark & \NoMark
\\

Topological Pooling~\citep{chen2023topologicalpooling}
&
\NoMark & \NoMark & \NoMark & \NoMark & \NoMark
& \Primary & \NoMark & \NoMark
\\

TopER~\citep{tola2025toper}
&
\NoMark & \NoMark & \NoMark & \NoMark & \NoMark
& \Primary & \NoMark & \NoMark
\\

Connectivity-Optimized RL~\citep{hofer2019connectivity}
&
\NoMark & \NoMark & \NoMark & \NoMark & \NoMark
& \Primary & \Primary & \NoMark
\\

TopoAE~\citep{moor2020topological}
&
\NoMark & \NoMark & \NoMark & \NoMark & \NoMark
& \Primary & \Primary & \NoMark
\\

PH-SSL Molecular Learning~\citep{luo2023improving}
&
\NoMark & \NoMark & \Primary & \NoMark & \NoMark
& \Primary & \Primary & \NoMark
\\


\midrule

\rowcolor{oursyellow}

\textbf{\textsc{Ours}}
&
\Primary
&
\Primary
&
\Primary
&
\Primary
&
\Primary
&
\Primary
&
\Primary
&

\textbf{\Primary}
\\

\bottomrule

\end{tabularx}

\vspace{3pt}

\begin{minipage}{\textwidth}
\scriptsize

\Primary\ = primary focus;
\Partial\ = partial or indirect consideration;
\NoMark\ = not addressed.

Meta. = metamaterials;
TPMS = triply periodic minimal surfaces;
TDA/PH = topological data analysis / persistent homology.

The blue sections denote general metamaterial and TPMS-oriented studies,
the green section denotes topology-aware studies,
and the blue--green rows indicate works belonging to both
TPMS-specific and topology-aware categories.
The yellow row highlights our method.

\end{minipage}

\end{center}

\end{table}

\begin{figure}[t]
    \centering
    \includegraphics[width=\linewidth]{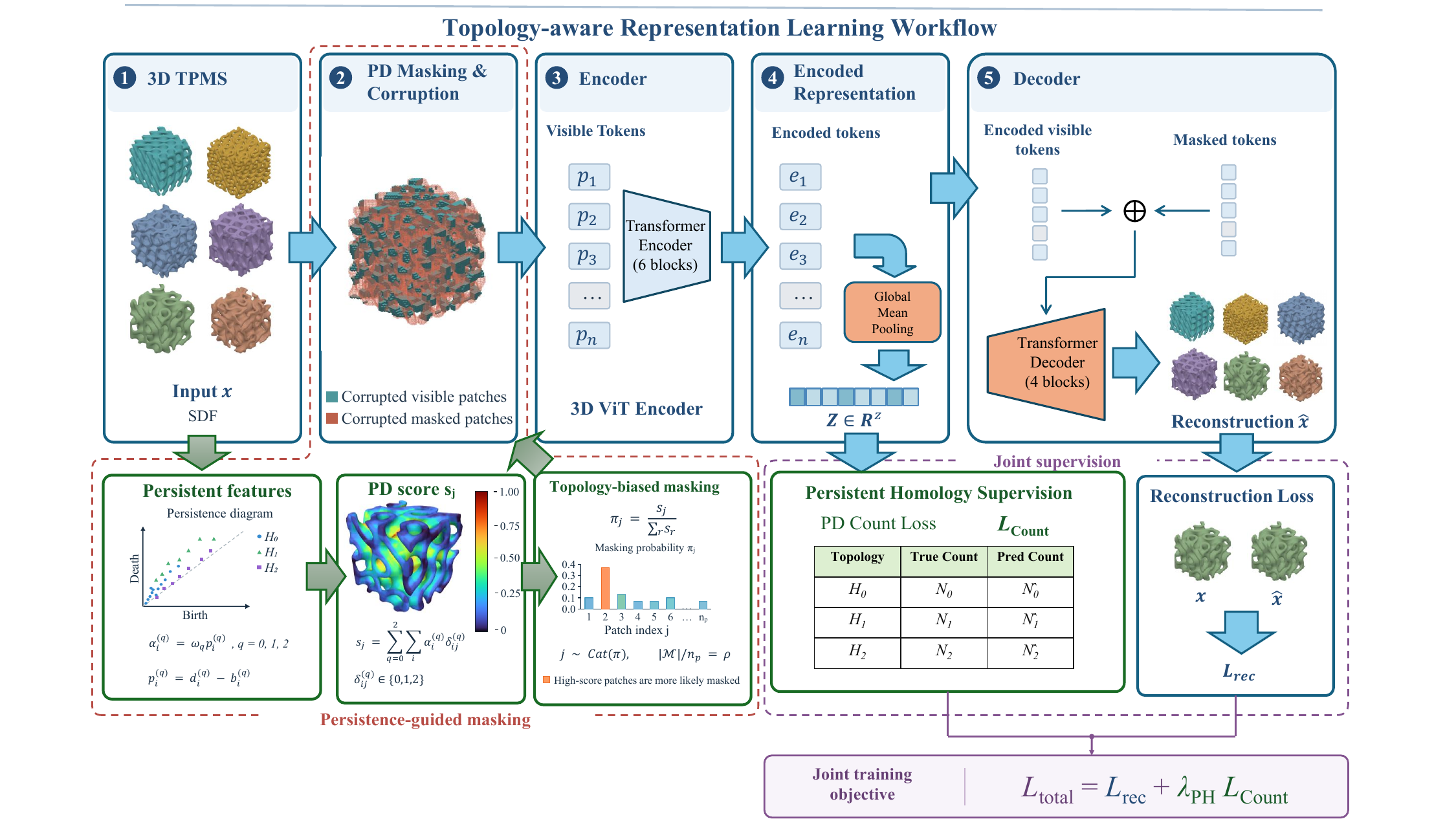}
    \caption{
    \textbf{Overview of the proposed topology-aware representation learning framework.}
    Starting from voxelized TPMS signed-distance fields, the framework integrates
    persistence-guided masking and a persistent-homology prediction objective
    into masked representation learning to preserve topology-relevant information
    in the latent representation.
    }
    \label{fig:workflow}
\end{figure}

Here, we introduce a topology-aware representation learning framework for mechanics-relevant TPMS metamaterials, as illustrated in Figure.~\ref{fig:workflow}. Rather than using topology only as an external descriptor or generative constraint, we incorporate persistent-homology information directly into representation learning through two complementary mechanisms. Inspired by masking-based self-supervised representation learning in MAE~\citep{he2022mae} and DMAE~\citep{wu2023dmae}, we introduce a \textbf{persistence-guided masking} strategy that prioritizes topologically informative regions, encouraging the encoder to integrate local geometry with global structural context. In parallel, an auxiliary \textbf{persistent-homology prediction head} explicitly supervises the latent representation to retain information about topological feature counts across connected components, tunnels, and voids. In this way, \textbf{guided masking determines where the model should focus during representation learning}, while the \textbf{PH head constrains how many topological features should be preserved in latent space}. We evaluate representations from geometric and mechanical perspectives and show that topology-aware learning produces more structurally and mechanically informative latent representations for downstream property analysis.

\raggedbottom

\section{Topology-aware representation learning workflow}
\label{sec:method}

Our framework extends a 3D DMAE~\citep{wu2023dmae} based on Vision Transformers~\citep{dosovitskiy2021image} by integrating persistent homology at the masking and latent-representation levels. Sec.~\ref{sec:patch_masking} describes patch-based SDF masking and reconstruction for TPMS. Sec.~\ref{sec:masking} introduces persistence-guided masking, distinct from existing random or adaptive schemes~\citep{he2022mae,hinojosa2024colormae,liu2024continualmae}. Sec.~\ref{sec:ph_supervision} presents auxiliary PH supervision that encourages the latent representation to retain global topological statistics. Architectural details are provided in Appendix~\ref{sec:model_architecture}.

\subsection{Patch Masking of TPMS SDFs}
\label{sec:patch_masking}

Given an TPMS SDF tensor $\tS\in\R^{N\times N\times N}$, we partition it into non-overlapping cubic patches of size $p\times p\times p$ using a patchification operator $\mathcal{P}(\cdot)$. Let $g=N/p$ denote the number of patches along each spatial dimension, yielding $n_p=g^3$ patches in total. The resulting clean patches are denoted by $\{\tP_j\}_{j=1}^{n_p}=\mathcal{P}(\tS)$, where $\tP_j\in\R^{p\times p\times p}$, and serve as reconstruction targets throughout training.

Following DMAE, Gaussian corruption is first applied to the clean SDF as $\widetilde{\tS}=\tS+\boldsymbol{\epsilon}$, where $\boldsymbol{\epsilon}$ denotes Gaussian noise, yielding the corresponding corrupted patches $\{\widetilde{\tP}_j\}_{j=1}^{n_p}=\mathcal{P}(\widetilde{\tS})$. For a masking ratio $\rho$, we sample a masked set $\sM\subset\{1,\ldots,n_p\}$ with $|\sM|\approx\rho n_p$, while $\sV=\{1,\ldots,n_p\}\setminus\sM$ denotes the visible/unmasked set. In our framework, $\sM$ is sampled according to the topology-guided masking distribution described in Sec.~\ref{sec:masking}. Only the corrupted visible patches $\{\widetilde{\tP}_j:j\in\sV\}$ are embedded and processed by the encoder $E_\theta$, producing latent tokens $\{\mathbf{h}_j\}_{j\in\sV}=E_\theta(\{\widetilde{\tP}_j\}_{j\in\sV})$.

For reconstruction, the encoded visible tokens $\mathbf{h}_j$ are combined with a learnable mask token $\mathbf{m}$ at positions $j\in\sM$ and restored to their original spatial ordering before being passed to the decoder $D_\phi$. The decoder therefore predicts all reconstructed patches as $\{\hat{\tP}_j\}_{j=1}^{n_p}=D_\phi(\{\mathbf{h}_j\}_{j\in\sV},\{\mathbf{m}\}_{j\in\sM})$. Following DMAE, reconstruction is evaluated over all $n_p$ patches against the clean targets $\tP_j$:
\begin{equation}
\mathcal{L}_{\mathrm{rec}}
=
\frac{1}{n_p}
\sum_{j=1}^{n_p}
\ell\!\left(\hat{\tP}_j,\tP_j\right),
\end{equation}
where $\ell$ denotes the Smooth L1 loss. For $j\in\sV$, the model denoises the corrupted observation $\widetilde{\tP}_j$ toward its clean counterpart $\tP_j$, whereas for $j\in\sM$, it reconstructs $\tP_j$ from the surrounding visible context. The combination of Gaussian corruption and topology-guided patch masking therefore encourages the encoder to learn representations that remain informative under both local perturbations and missing structural observations.

\subsection{Persistence-Guided Topological Masking}
\label{sec:masking}

Existing MAE-based methods have explored different masking strategies, including uniform random masking in vanilla MAE~\citep{he2022mae}, filtered-noise data-independent masking in ColorMAE~\citep{hinojosa2024colormae}, and distribution-aware adaptive masking in DaM~\citep{liu2024continualmae}. 
While these strategies modify the spatial distribution of masked patches, they do not explicitly account for persistent topological information. 
We instead construct a patch-level topological importance score from persistent homology and use it to bias the masking distribution toward topology-relevant regions. 
Details of the persistent-homology computation and its formulation for TPMS structures are provided in Appendix~\ref{sec:PH_TPMS}.

\noindent
\colorbox{gray!10}{
\parbox{\dimexpr\linewidth-2\fboxsep\relax}{
\textbf{Insights:}
(1) As shown in Fig.~\ref{fig:cut_merge} and Appendix~\ref{sec:topo_sensitivity}, localized topology-changing perturbations can modify structural connectivity and load-transfer pathways, leading to noticeable changes in effective stiffness even when the geometric variation is small. These changes are also reflected in the corresponding persistence diagrams. For example, in the \textit{Cut} case, the perturbation alters the persistence of topological features, coinciding with a clear change in the elastic response, which suggests that persistent homology captures structural variations relevant to mechanical behavior. (2) Within a persistence diagram, features with larger persistence remain present over a wider range of filtration scales and therefore represent more robust topological characteristics~\citep{edelsbrunner2002topological,zomorodian2005computing}.

}
}

\noindent
\colorbox{gray!10}{
\parbox{\dimexpr\linewidth-2\fboxsep\relax}{
\textbf{Assumption:}
We therefore hypothesize that preferentially masking regions associated with highly persistent topological features encourages the model to better learn and reconstruct mechanically relevant structural information. We therefore assign each patch a persistence-weighted topological score and use it to define the masking probability.
}
}

Specifically, the topological importance score of patch \(j\) is defined as
\(s_j=\sum_{q=0}^{2}\sum_i \alpha_i^{(q)}\delta_{ij}^{(q)}\),
where \(\alpha_i^{(q)}=\omega_q p_i^{(q)}\), \(p_i^{(q)}\) denotes the persistence of the \(i\)-th topological feature in homology dimension \(q\), and \(\omega_q\) controls the relative contribution of dimension \(q\).
The term \(\delta_{ij}^{(q)}\in\{0,1,2\}\) counts how many of the birth and death critical cells associated with feature \(i\) are mapped to patch \(j\).
Thus, features with larger persistence contribute more strongly to the score, while \(\omega_q\) allows different homology dimensions to be weighted separately.
Accordingly, a patch receives a larger score when it contains critical cells associated with more persistent topological features.

The patch scores are then normalized into masking probabilities,
\(\pi_j=s_j/\sum_{r=1}^{n_p}s_r\), for \(j\in\{1,\ldots,n_p\}\).
Thus, patches with larger \(\pi_j\) are more likely to be selected for masking and are sampled from this distribution until the prescribed masking ratio \(\rho\) is reached.
Representative PD-score distributions for the baseline structure and four localized topological modifications are visualized in Appendix Fig.~\ref{fig:pd_score_maps}.
Additional PD-score heatmaps for diverse TPMS structures are provided in Appendix~\ref{sec:PH_diversity} and Fig.~\ref{fig:PH_diversity}.

\subsection{Persistent-Homology Supervision}
\label{sec:ph_supervision}

We introduce an auxiliary PH prediction head operating on the pooled encoder representation.
Since the number of retained features can vary substantially across structures, we use log-transformed feature counts as the prediction targets, with
\(\hat{\mathbf{c}}=[\hat c_0,\hat c_1,\hat c_2]\).
The PH supervision loss is defined as,

\begin{equation}
\mathcal{L}_{\mathrm{PH}}
=
\sum_{q=0}^{2}
\ell_{\mathrm{SL1}}
\left(
\hat c_q,\log(1+n_q)
\right),
\end{equation}
where \(n_q\) denotes the number of retained persistence features in homology dimension \(q\in\{0,1,2\}\), \(\log(1+n_q)\) is the corresponding log-transformed count target, and \(\ell_{\mathrm{SL1}}\) denotes the Smooth L1 loss. The logarithmic transformation reduces the dynamic range of the feature counts and stabilizes regression across structures with substantially different topological complexity.

\noindent
\colorbox{gray!10}{
\parbox{\dimexpr\linewidth-2\fboxsep\relax}{
\textbf{Insights:}
For Fig.~\ref{fig:cut_merge} and Table.~\ref{tab:perturbation_ph_counts}, the \textit{Merged} modification changes the $H_1$ and $H_2$ feature counts by $-203$ and $-35$, respectively, while also producing pronounced changes in the elastic response. 
These count variations reflect changes in global topological complexity that are not explicitly supervised by the persistence-guided spatial masking. 
}
}

\noindent
\colorbox{gray!10}{
\parbox{\dimexpr\linewidth-2\fboxsep\relax}{
\textbf{Assumption:}
We therefore hypothesize that explicitly requiring the latent representation to predict global $H_0$, $H_1$, and $H_2$ feature counts can complement persistence-guided masking and encourage mechanics-relevant global topological information to be retained after representation compression.
}
}

The complete training objective is
\begin{equation}
\mathcal{L}_{\mathrm{total}}
=
\mathcal{L}_{\mathrm{rec}}
+
\lambda_{\mathrm{PD}}\mathcal{L}_{\mathrm{PH}},
\end{equation}

where \(\lambda_{\mathrm{PD}}\) controls the contribution of PH supervision to the overall training objective.

\section{Experiments}
\label{sec:experiments}

To evaluate the assumptions introduced in Sec.~\ref{sec:method}, we assess the learned representations through downstream tasks, ablation studies, and latent-space analysis.
We first describe the construction of the two TPMS datasets in Sec.~\ref{sec:dataset} and the competing baselines in Sec.~\ref{sec:baselines}.
Sec.~\ref{sec:downstream} evaluates structural and mechanics-relevant information through four downstream tasks, while Sec.~\ref{sec:ablation} examines the effects of persistence-guided masking and PH supervision.
Finally, Sec.~\ref{sec:latent_alignment} analyzes how the learned latent space aligns with topology- and mechanics-related descriptors.

\subsection{Dataset}
\label{sec:dataset}
We used two complementary TPMS datasets to evaluate representation learning across procedurally generated and continuously interpolated structural design spaces. 

\textbf{MetaDB.} The open-source MetaDB dataset represents metamaterial geometries using the MetaDSL procedural representation and expands hand-authored structures through type-aware mutation operators~\citep{makatura2025metagen}. From MetaDB, we select 13 base TPMS types and their mutation-derived variants, resulting in 13,114 structures. In the original MetaDB, each geometry is voxelized and its effective elastic properties are obtained through periodic homogenization using the finite-element framework ProcMeta~\citep{makatura2025metagen}. ProcMeta assumes an isotropic base material with normalized Young's modulus $E=1$, density $\rho=1$, and Poisson's ratio $\nu=0.45$, and provides the resulting homogenized stiffness tensor $\mathbf{C}\in\mathbb{R}^{6\times6}$ in Voigt notation as the mechanical-property label. Further details on the stiffness tensor and the finite-element homogenization procedure are provided in Appendix~\ref{sec:fe_homogenization}. We resampled all voxelized geometries to a resolution of $64^3$.

\textbf{Interpolation.} We constructed a second dataset based on ~\citep{perronischarf2025tpms} by interpolating eight canonical TPMS primitives. Each primitive was represented by a periodic implicit function $f_i(x,y,z)$, and the interpolated geometry was defined as $F(x,y,z)=\sum_{i=1}^{8} w_i f_i(x,y,z)$.

We obtained the interpolated TPMS geometry from the zero-level set $F(x,y,z)=0$. For each interpolated geometry, we combined up to six base TPMS families, with each nonzero interpolation weight taking values from $10\%$ to $100\%$ in increments of $10\%$, while satisfying $\sum_{i=1}^{8} w_i = 1$. Using this interpolation scheme, we generated 18,549 structures and represented each geometry as a $64^3$ SDF. We then applied the same periodic homogenization framework, ProcMeta, used for MetaDB to compute the corresponding $6\times6$ effective stiffness tensor $\mathbf{C}$ as the mechanical-property label.

\subsection{Baselines}
\label{sec:baselines}
We compare our proposed method with geometry-based representation-learning baselines. 
For general representation learning, we include MVAE~\citep{choy2019minkowski}, 
SVAE~\citep{xiang2025structured}, 
VICReg~\citep{bardes2022vicreg}, and 
I-JEPA~\citep{assran2023ijepa}. 
For masking-strategy baselines, we use 
MAE~\citep{he2022mae}, 
DMAE~\citep{wu2023dmae}, 
ColorMAE~\cite{hinojosa2024colormae}, and 
DaM~\cite{liu2024continualmae}. 
Specifically, MAE uses random patch masking, whereas DMAE combines random masking with Gaussian input corruption. 
ColorMAE generates structured masks by filtering random noise, while DaM adaptively selects masked tokens according to estimated distribution shifts. 
Additionally, we include ScalableTopo~\citep{wong2025scalabletopo} as a topology-informed baseline based on scalable persistent-homology regularization.

\subsection{Evaluation on Downstream Tasks}
\label{sec:downstream}
We evaluate latent representations on four downstream tasks: family classification, mechanical property classification, clustering, and regression. The first probes TPMS structural identity, while the others assess the encoding of effective mechanical properties. All experiments use a fixed 70/15/15 train/validation/test split, with results reported over five random seeds. Models are trained and evaluated on NVIDIA H100 GPUs on Compute Canada clusters.

\subsubsection{Family Classification}
We evaluate frozen representations using a linear probe and a $k$-nearest-neighbor ($k$NN) classifier ($k=15$). For MetaDB, the 13 base TPMS families serve as single-label targets, with macro-F1 reported as $\mathrm{F}_{\mathrm{Lin}}$ and $\mathrm{F}_{k\mathrm{NN}}$, respectively. For Interpolation TPMS, labels correspond to families with the largest interpolation weight, including ties. Thus, $50\%$ Gyroid--$30\%$ FRD--$20\%$ PMY is labeled Gyroid, whereas $50\%$ Gyroid--$50\%$ Diamond receives both labels. Strict accuracy ($\mathrm{Acc}^{\mathrm{Str}}$) requires exact agreement between predicted and true label sets; relaxed accuracy ($\mathrm{Acc}^{\mathrm{Rel}}$) accepts any nonempty predicted subset of the true labels.

As shown in Table~\ref{tab:family_classification}, our method achieves the best performance on MetaDB for both linear and $k$NN probes. On Interpolation TPMS, it achieves the highest $k$NN accuracy under both strict and relaxed evaluation, while the linear-probe results remain slightly below the best baseline. These results indicate that topology-aware learning preserves family-relevant structural information while promoting a more coherent local organization of structurally similar samples in the latent space.

\begin{table*}[t]
\caption{
\textbf{Task 1 -- Family classification.}
$\mathrm{F}_{\mathrm{Lin}}$ and $\mathrm{F}_{k\mathrm{NN}}$ denote macro-F1 scores on MetaDB using a linear probe and $k$NN classifier, respectively.
For Interpolation TPMS, $\mathrm{Acc}_{\mathrm{Lin}}^{\mathrm{Str/Rel}}$ and $\mathrm{Acc}_{k\mathrm{NN}}^{\mathrm{Str/Rel}}$ denote strict and relaxed accuracies with respect to the dominant-family labels.
Higher is better for all metrics.
}
\label{tab:family_classification}

\begin{center}
\small
\setlength{\tabcolsep}{5pt}
\renewcommand{\arraystretch}{1.08}

\begin{tabular}{lcc|cccc}
\hline

\textsc{Model}
& \multicolumn{2}{c|}{\textsc{MetaDB}}
& \multicolumn{4}{c}{\textsc{Interpolation TPMS}}
\\

\cline{2-3}
\cline{4-7}

&
$\mathrm{F}_{\mathrm{Lin}}\uparrow$
&
$\mathrm{F}_{k\mathrm{NN}}\uparrow$
&
$\mathrm{Acc}_{\mathrm{Lin}}^{\mathrm{Str}}\uparrow$
&
$\mathrm{Acc}_{\mathrm{Lin}}^{\mathrm{Rel}}\uparrow$
&
$\mathrm{Acc}_{k\mathrm{NN}}^{\mathrm{Str}}\uparrow$
&
$\mathrm{Acc}_{k\mathrm{NN}}^{\mathrm{Rel}}\uparrow$
\\

\hline

\multicolumn{7}{l}{\textbf{\textit{General representation learning}}}
\\

MVAE 
& 0.009
& 0.581
& 0.038
& 0.051
& 0.446
& 0.559
\\

SVAE 
& 0.479
& 0.955
& 0.638
& 0.663
& 0.787
& 0.914
\\

VICReg 
& 0.913
& 0.853
& 0.263
& 0.299
& 0.370
& 0.456
\\

I-JEPA 
& 0.912
& 0.902
& 0.403
& 0.446
& 0.497
& 0.618
\\

\hline

\multicolumn{7}{l}{\textbf{\textit{Masking strategy baselines}}}
\\

MAE 
& 0.901
& 0.907
& 0.549
& 0.588
& 0.643
& 0.792
\\

DMAE 
& 0.907
& 0.894
& 0.724
& 0.852
& 0.819
& 0.932
\\

ColorMAE 
& 0.956
& 0.952
& 0.877
& 0.888
& 0.821
& 0.931
\\

DaM 
& 0.952
& 0.948
& 0.879
& 0.893
& 0.829
& 0.937
\\

\hline

\multicolumn{7}{l}{\textbf{\textit{Topology-informed baseline}}}
\\

ScalableTopo 
& 0.957
& 0.954
& \textbf{0.885}
& \textbf{0.897}
& 0.813
& 0.928
\\

\hline

\textsc{Ours}
& \textbf{0.994}
& \textbf{0.985}
& 0.867
& 0.880
& \textbf{0.831}
& \textbf{0.939}
\\

\hline

\end{tabular}
\end{center}
\end{table*}

\subsubsection{Mechanical-property prediction}

\begin{figure*}[!t]
    \centering
    \includegraphics[width=0.7\linewidth]{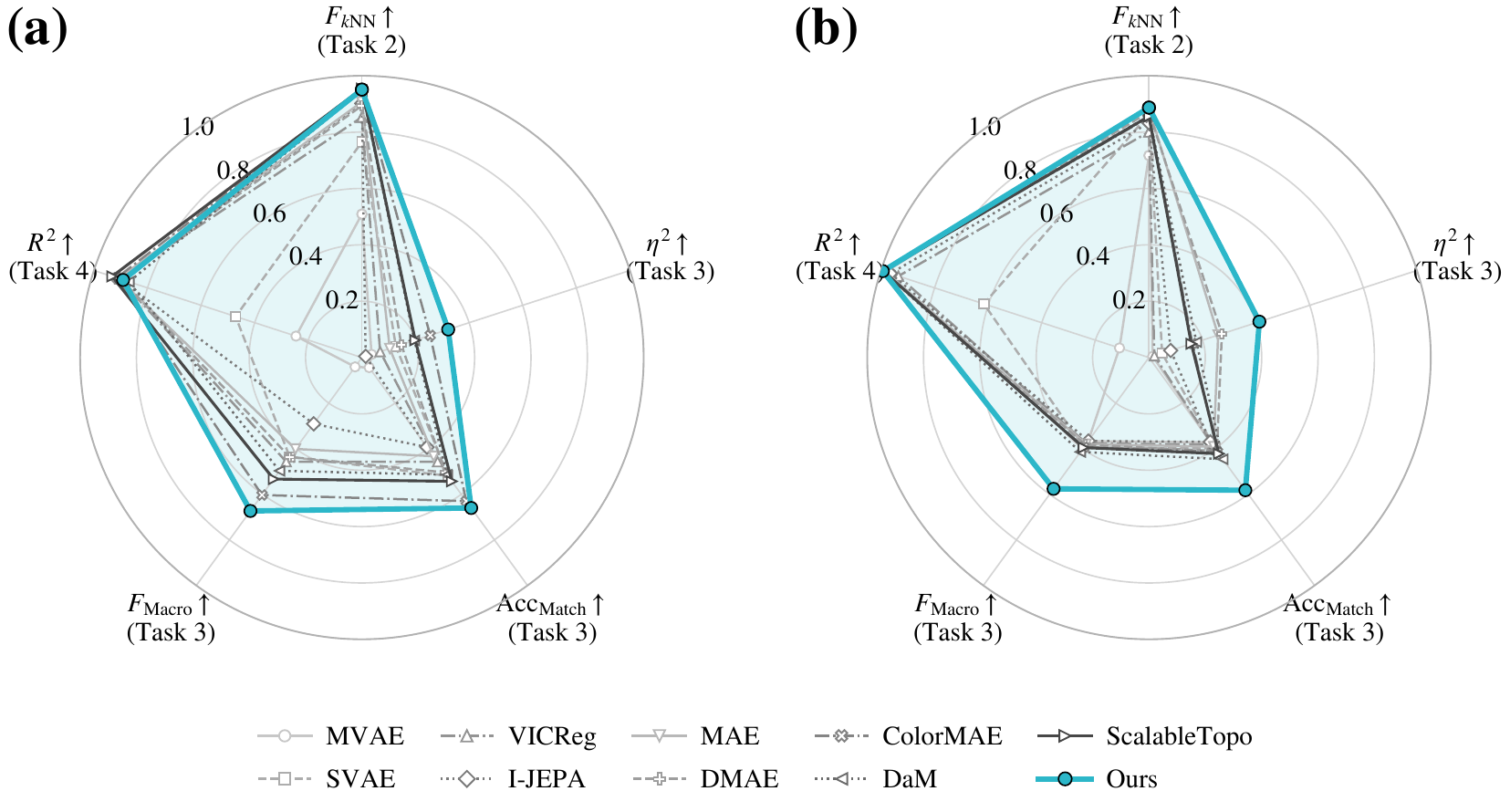}
    \caption{
    \textbf{Summary of mechanical-property downstream performance on (a) MetaDB and (b) Interpolation TPMS.}
    The radar plots provide a compact comparison of frozen latent representations: 
    $\mathrm{F1}_{k\mathrm{NN}}$ for $C_{11}$ property classification (Task 2), 
    $\eta^2$, $\mathrm{Acc}_{\mathrm{Match}}$, and $\mathrm{F}_{\mathrm{Macro}}$ for $C_{11}$ property clustering (Task 3), 
    and $R^2$ for $C_{11}$ property regression (Task 4). 
    Higher is better for all metrics shown. 
    Detailed numerical results for all metrics and baselines are provided in the Appendix 
    (Tables~\ref{tab:c11_classification}, \ref{tab:c11_clustering}, and~\ref{tab:c11_regression}).
    }
    \label{fig:c11_summary}
\end{figure*}

We evaluate mechanics-relevant information in the frozen representations through three downstream tasks. $C_{11}$, the effective normal stiffness along a principal direction, is used for property classification (Task 2), clustering (Task 3), and regression (Task 4). Figure~\ref{fig:c11_summary} summarizes representative results across both datasets, with complete numerical results provided in Appendix~\ref{sec:downstream_appendix}.

\paragraph{Task 2 -- $C_{11}$ property classification.}
We discretize $C_{11}$ into \textit{Low}, \textit{Medium}, and \textit{High} groups and evaluate linear and $k$NN ($k=15$) probes using balanced accuracy and macro-F1. Figure~\ref{fig:c11_summary} highlights $k$NN macro-F1 ($F_{k\mathrm{NN}}$): our method achieves the highest score on Interpolation TPMS (0.887) and remains close to the best baseline on MetaDB (0.951 versus 0.954). The linear probe further achieves the highest balanced accuracy and macro-F1 on both datasets, supporting effective discrimination between stiffness groups.

\paragraph{Task 3 -- $C_{11}$ property clustering.}
We assess whether the latent space organizes structures according to $C_{11}$ without supervision using $\eta^2$, $\mathrm{Acc}_{\mathrm{Match}}$, and $F_{\mathrm{Macro}}$. As shown in Fig.~\ref{fig:c11_summary}, our method achieves the highest values across all three metrics on both datasets. Relative to the strongest competing baseline, it improves the three metrics by $26.7\%$, $4.8\%$, and $11.6\%$ on MetaDB, and by $51.5\%$, $30.5\%$, and $39.1\%$ on Interpolation TPMS, respectively. These results demonstrate a substantially stronger mechanics-aware organization of the latent space.

\paragraph{Task 4 -- $C_{11}$ property regression.}
Finally, we assess whether continuous mechanical-property information can be predicted from the frozen representations through $C_{11}$ regression. Our method achieves $R^2=0.899$ on MetaDB and $R^2=0.993$ on Interpolation TPMS, showing the best performance on the latter while maintaining low prediction errors across both datasets. These results show that the learned representations retain continuous mechanics-relevant information in addition to discrete property organization.

Overall, Tasks 2--4 show that our method consistently preserves mechanics-relevant information, with particularly strong improvements in unsupervised property clustering while maintaining competitive classification and regression performance.

\subsection{Ablation Study}
\label{sec:ablation}

\begin{figure*}[t]
    \centering
    \includegraphics[width=0.8\textwidth]{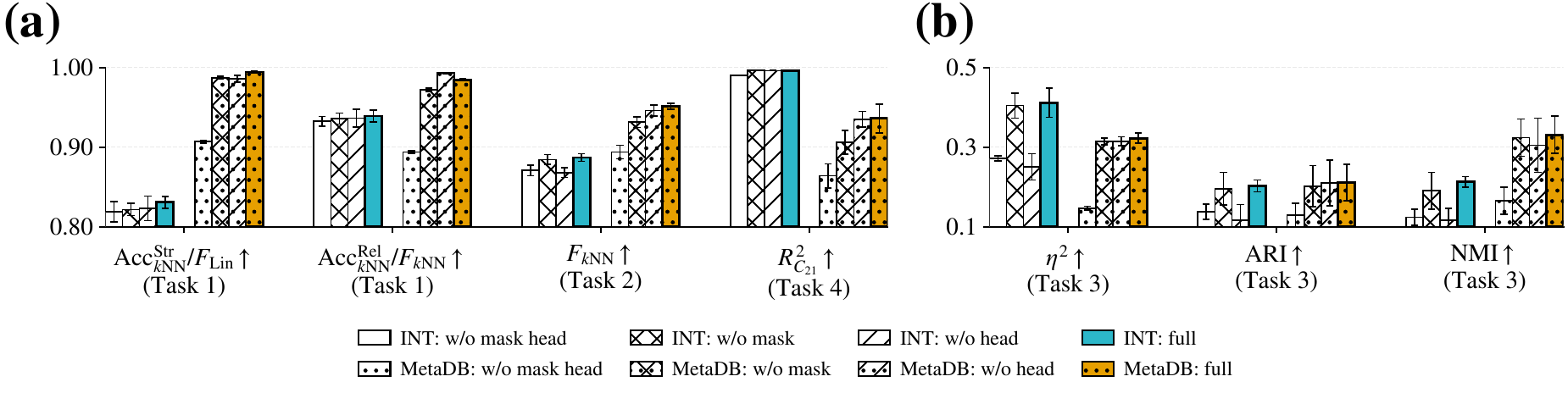}
    \caption{
    \textbf{Ablation study of persistence-guided masking and the auxiliary PD prediction head.} \textbf{(a)} reports representative metrics from Tasks~1, 2, and 4, while \textbf{(b)} reports the three Task~3 clustering metrics. We compare the DMAE baseline without either topology-aware mechanism (\textit{w/o mask head}), variants removing either the persistence-guided mask (\textit{w/o mask}) or PD head (\textit{w/o head}), and the full model using both mechanisms (\textit{full}). Complete numerical results across all evaluation metrics and both datasets are provided in Appendix Table~\ref{tab:ablation_full}. Bars show the mean over five seeds, with error bars indicating the corresponding standard deviation. Higher is better for all metrics.
    }
    \label{fig:ablation}
\end{figure*}

Figure~\ref{fig:ablation} examines the individual and combined effects of persistence-guided masking and the auxiliary PD prediction head across both datasets. The full model provides the most consistent overall high performance and achieves the strongest Task~3 clustering results in terms of $\eta^2$, ARI, and NMI on both datasets, indicating improved organization of mechanics-relevant information in the latent space. The additional \(C_{21}\) (with 21 independent elastic stiffness components in Voigt notation) evaluation shows a similar trend, indicating that the benefit is not restricted to a single elastic-property target. Although individual ablations can perform better on isolated metrics, neither component alone provides the same overall consistency. Definitions of ARI and NMI used for the clustering evaluation are provided in Appendix Sec.~\ref{sec:clustering_metrics}.

\subsection{Latent-Space Alignment Analysis}
\label{sec:latent_alignment}

\begin{figure}[t]
    \centering
    \includegraphics[width=0.8\linewidth]{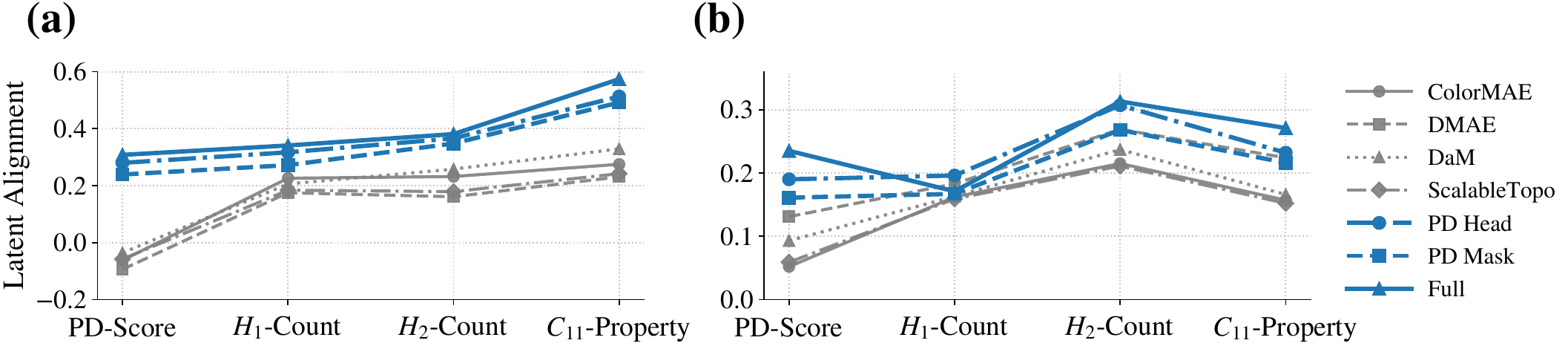}
    \caption{
    \textbf{Direct latent-space alignment on (a) MetaDB and (b) Interpolation TPMS.} PD-Score, $H_1$-Count, $H_2$-Count, and $C_{11}$-Property measure the correspondence between pairwise latent distances and the persistence-guided masking distribution, persistence-feature counts, and mechanical-property differences, respectively. Higher Spearman correlation indicates stronger alignment.
    }

    \label{fig:latent_alignment}
\end{figure}

To directly assess the learned latent space, we measure the Spearman correlation between pairwise latent distances and four descriptors: the PD-score masking distribution, $H_1$ and $H_2$ persistence-feature counts, and $C_{11}$. These metrics evaluate whether the latent space reflects the spatial topology distribution used for PD masking, the topological statistics supervised by the PD head, and mechanics-related variation. Detailed definitions of the alignment metrics are provided in Appendix~\ref{sec:latent_alignment_metrics}. As shown in Fig.~\ref{fig:latent_alignment}, topology-aware learning generally strengthens these latent correspondences across both datasets. On MetaDB, the full model achieves the strongest alignment for all four metrics. On Interpolation TPMS, the full model performs best for PD-score, $H_2$-count, and $C_{11}$ alignment, while the PD head achieves the strongest $H_1$-count alignment. These results indicate complementary effects of topology-guided masking and explicit topological supervision.

\section{Conclusion}

We presented a topology-aware representation learning framework for TPMS metamaterials that explicitly integrates persistent homology into DMAE. The proposed method combines persistence-guided masking, which directs learning toward topologically informative structural regions, with an auxiliary PH prediction head that encourages the latent representation to retain topological information. The learned representations demonstrate strong performance in structural family classification and multiple mechanical-property downstream tasks, including $C_{11}$ classification, clustering, and regression. In particular, the results show improved organization of mechanically meaningful latent structure, especially for continuously interpolated TPMS geometries. Ablation studies reveal complementary roles of the two topology-aware components, while direct latent-space analysis shows stronger correspondence with PD-score distributions, persistence-feature counts, and mechanical properties.  Overall, these results demonstrate that persistent topology provides a useful inductive bias for learning compact representations of complex three-dimensional architected materials and provide a foundation for future topology-aware property prediction and generative design.

\clearpage
\subsection*{AI Use Statement}

In this work, we used generative AI tools for manuscript polishing as well as code debugging. We have not used
generative AI tools for developing the research ideas, methodology, experimental design, or scientific claims, nor to generate datasets or mathematical results. All AI-assisted code modifications were reviewed, executed, and verified by two authors, and all AI-assisted text was carefully reviewed and revised before inclusion in the manuscript. We take full responsibility for the final content of this work, including any text or code produced with the aid of generative AI.

\subsection*{Reproducibility Statement}

To facilitate reproducibility, we provide detailed implementation and evaluation information in the supplementary material.
Appendix~\ref{sec:topo_sensitivity} presents the controlled topology--mechanics perturbation analysis and persistence-guided masking behavior; Sec.~\ref{sec:model_architecture} specifies the model architecture, patchification, encoder--decoder configuration, and training objectives; Sec.~\ref{sec:PH_TPMS} describes the persistent-homology pipeline, including the SDF-based cubical filtration and persistence-diagram construction for TPMS; Sec.~\ref{sec:PH_diversity} illustrates the topological diversity of the dataset; Sec.~\ref{sec:fe_homogenization} details the finite-element homogenization procedure used to obtain the mechanical-property labels; and Sec.~\ref{sec:downstream_appendix} provides the downstream evaluation protocols, metric definitions, and complete results for Tasks~2--4.
Complete ablation results are reported in Table~\ref{tab:ablation_full}, with the additional clustering metrics used in the ablation analysis defined in Sec.~\ref{sec:clustering_metrics}. Sec.~\ref{sec:latent_alignment_metrics} provides the definitions of the latent-space alignment measures used in the representation analysis, and Sec.~\ref{sec:param_sensitivity} reports parameter-sensitivity analyses and hyperparameter selection on both Interpolation TPMS and MetaDB.

The MetaDB dataset used in this study is derived from a publicly available open-source dataset, while the Interpolation TPMS dataset constructed for this work will be released on GitHub.
The source code and implementation details for the proposed method will also be released through the accompanying GitHub repository.

\bibliography{iclr2027_conference}
\bibliographystyle{iclr2027_conference}

\appendix
\section{Appendix}
\setcounter{table}{0}
\renewcommand{\thetable}{S\arabic{table}}
\setcounter{figure}{0}
\renewcommand{\thefigure}{S\arabic{figure}}

\subsection{Topological Sensitivity Analysis}
\label{sec:topo_sensitivity}

To further examine the relationship between localized structural perturbations, persistent topology, and mechanical response, we construct a controlled experiment consisting of one baseline TPMS structure and four locally modified variants. Because the modifications are spatially confined while the overall geometry remains largely unchanged, these examples provide a controlled setting for studying how local connectivity changes are reflected in persistent-homology diagrams, elastic mechanical responses, and the persistence-guided masking heatmaps.

\begin{figure}[H]
    \centering
    \includegraphics[width=0.9\textwidth]{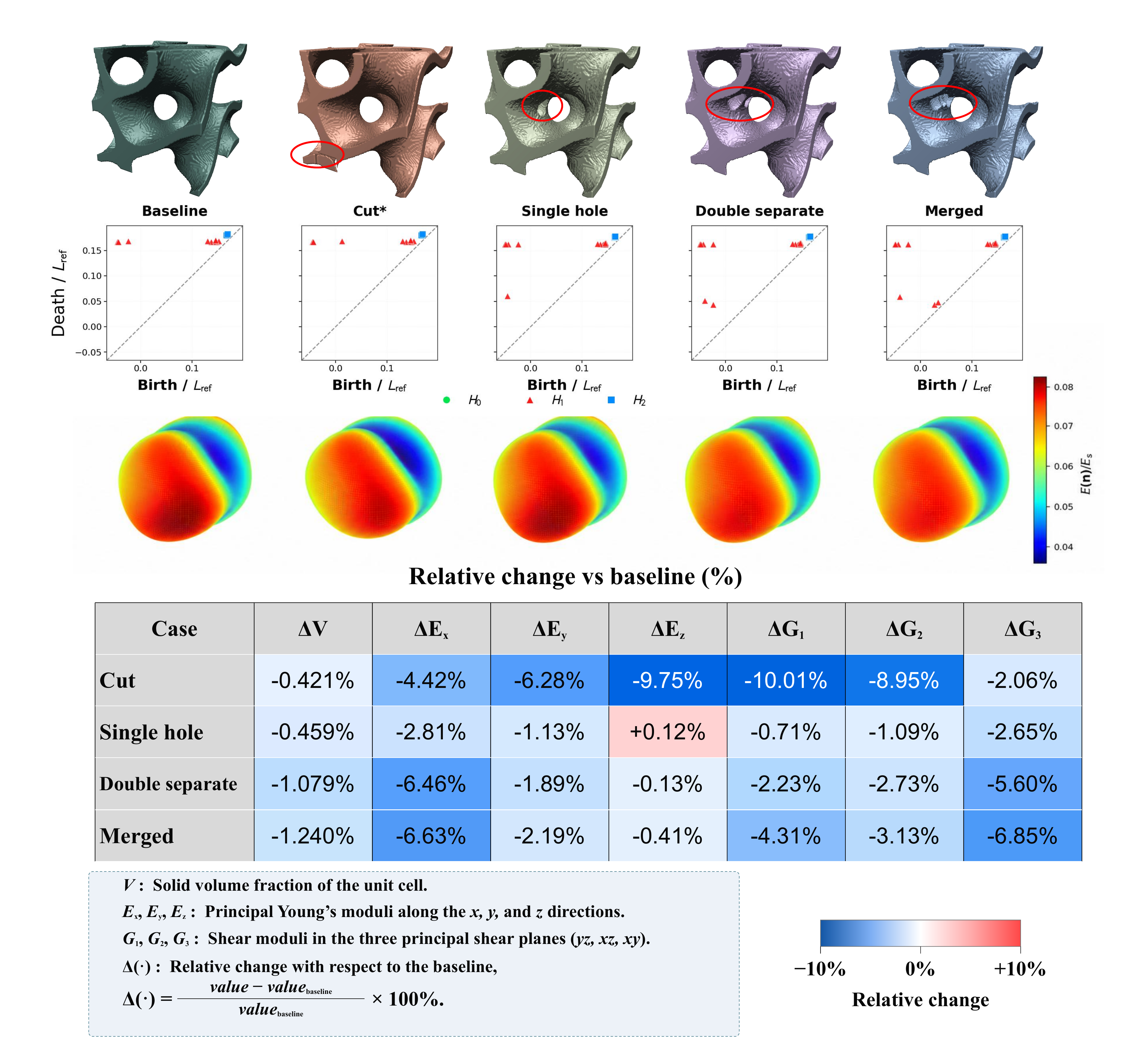}
    \caption{
    \textbf{Topological perturbations and their mechanical responses.} Five geometries are compared: the baseline structure and four locally modified variants (Cut, Single hole, Double separate, and Merged). The top row shows the corresponding geometries and localized modifications, the second row presents their persistence diagrams for $H_0$, $H_1$, and $H_2$, and the third row shows the directional Young's modulus surfaces. The table summarizes the relative changes in solid volume fraction and the six principal elastic moduli with respect to the baseline.
    }
    \label{fig:topology_mechanics}
\end{figure}

Figure~\ref{fig:topology_mechanics} compares four distinct localized modifications derived from the same baseline geometry. \textit{Cut} induces a crack to an existing connectivity pathway; \textit{Single hole} introduces one additional local opening; \textit{Double separate} introduces two nearby but disconnected openings; and \textit{Merged} connects the neighboring openings into a larger merged feature. These perturbations remain spatially localized, enabling comparison of structures with relatively small geometric changes but different alterations to local connectivity.

The resulting mechanical changes are not explained by the amount of removed material alone. The Cut and Single-hole modifications, for example, exhibit similar changes in solid volume fraction, $-0.421\%$ and $-0.459\%$, respectively, but substantially different elastic responses. The Cut structure produces reductions of $4.42\%$, $6.28\%$, and $9.75\%$ in $E_x$, $E_y$, and $E_z$, respectively, together with reductions of $10.01\%$, $8.95\%$, and $2.06\%$ in the corresponding shear moduli. In contrast, the The Single-hole modification leads to somewhat smaller, but still appreciable, mechanical changes relative to its small reduction in material volume, with $E_x$ and $E_y$ decreasing by $2.81\%$ and $1.13\%$, respectively, while $E_z$ remains nearly unchanged at $+0.12\%$. Thus, even for comparable and sub-percent changes in solid volume fraction, the mechanical response can vary noticeably depending on where and how the local structural connectivity is modified.

The Double-separate and Merged modifications produce larger changes in solid volume fraction, $-1.079\%$ and $-1.240\%$, respectively, and both exhibit larger reductions along selected directions. For Double separate, $E_x$ decreases by $6.46\%$ and the largest shear-modulus reduction reaches $5.60\%$. The Merged modification shows reductions of $6.63\%$ in $E_x$ and $6.85\%$ in $G_{xy}$. Although the two cases have comparable overall volume changes and visually similar local modifications, their directional elastic responses are not identical, indicating that the spatial organization and connectivity of the perturbation influence the resulting anisotropic stiffness.

These mechanical changes are accompanied by corresponding differences in the persistence diagrams (explained in Appendix~\ref{sec:PH_TPMS}) in Fig.~\ref{fig:topology_mechanics}. In particular, the localized modifications alter persistent $H_1$ and $H_2$ features associated with tunnels and voids, respectively. The comparison therefore shows that relatively small geometric perturbations can simultaneously modify persistent-topological descriptors and directional elastic behavior, motivating the use of persistent homology as an additional structural signal during representation learning.

\begin{figure}[H]
    \centering
    \includegraphics[width=0.9\textwidth]{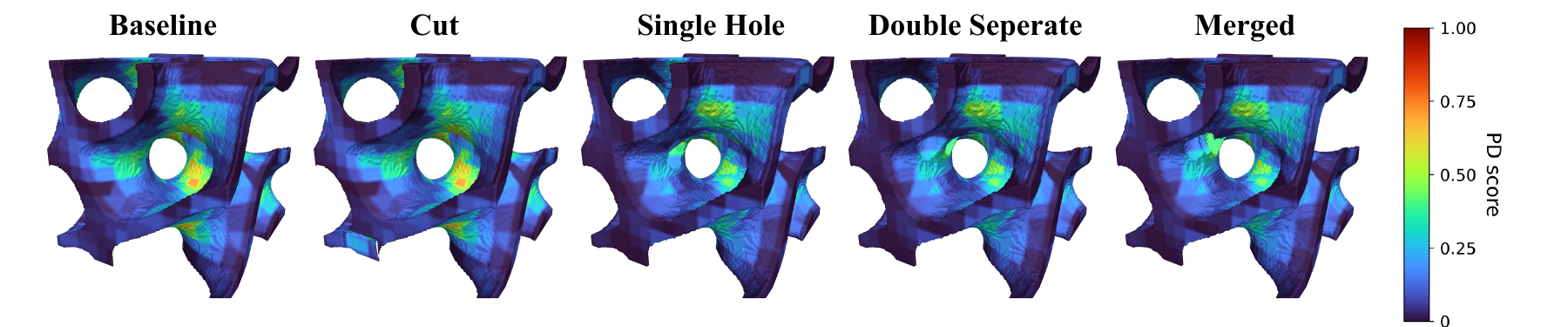}
    \caption{
    \textbf{Persistence masking-guided PD-score distributions under localized topological perturbations.}
    Normalized PD scores are projected onto the baseline structure and four locally modified variants. 
    Warmer colors indicate regions with larger PD scores and therefore higher probabilities of being selected by the persistence-guided masking strategy.
    }
    \label{fig:pd_score_maps}
\end{figure}

Figure~\ref{fig:pd_score_maps} further examines how the persistence-guided masking signal (key mechanism of the proposed model, described in Sec.~\ref{sec:masking}) responds to these perturbations. For the Cut modification, the PD-score distribution changes most noticeably around the interrupted region, whereas the Single-hole and Double-separate modifications exhibit score variations around the newly introduced openings. These results indicate that the masking distribution is sensitive to localized regions in which structural connectivity is altered.

The PD-score maps of the Double-separate and Merged modifications remain relatively similar despite distinguishable differences in their persistence diagrams in Fig.~\ref{fig:topology_mechanics}. This observation suggests that the spatial masking signal alone may not capture all persistent-topological differences, particularly when structures exhibit similar local importance patterns but differ in their global PH descriptors. The auxiliary PH prediction head therefore provides a complementary learning signal by encouraging PH-derived information to remain decodable from the latent representation. Together, these observations motivate the coordinated use of persistence-guided masking and PH-based latent supervision in the proposed framework.

\begin{table}[t]
\centering
\caption{
\textbf{Persistent-homology feature counts for the controlled topological perturbations.}
The table reports the occupied voxel count, voxel occupancy, and the numbers of finite persistence features in $H_0$, $H_1$, and $H_2$.
Values in parentheses denote the change relative to the baseline.
All structures contain one infinite $H_0$ feature and no infinite $H_1$ or $H_2$ features.
}
\label{tab:perturbation_ph_counts}

\small
\setlength{\tabcolsep}{5pt}
\renewcommand{\arraystretch}{1.10}

\begin{tabular}{lccccc}
\hline
\textbf{Structure}
& \textbf{Occupied voxels}
& \textbf{Occupancy (\%)}
& $\mathbf{H_0}$
& $\mathbf{H_1}$
& $\mathbf{H_2}$ \\
\hline

Baseline
& 162,042
& 18.315
& 373
& 2190
& 1150 \\

Cut
& 161,520
& 18.256
& 370 ($-3$)
& 2194 ($+4$)
& 1153 ($+3$) \\

Single hole
& 181,533
& 20.518
& 503 ($+130$)
& 2012 ($-178$)
& 1111 ($-39$) \\

Double separate
& 180,767
& 20.432
& 498 ($+125$)
& 1982 ($-208$)
& 1111 ($-39$) \\

Merged
& 180,465
& 20.398
& 498 ($+125$)
& 1987 ($-203$)
& 1115 ($-35$) \\

\hline
\end{tabular}
\end{table}

\subsection{Background on TPMS Shell Metamaterials}
\label{sec:tpms_shell_background}

\begin{figure}[H]
    \centering
    \includegraphics[width=0.8\textwidth]{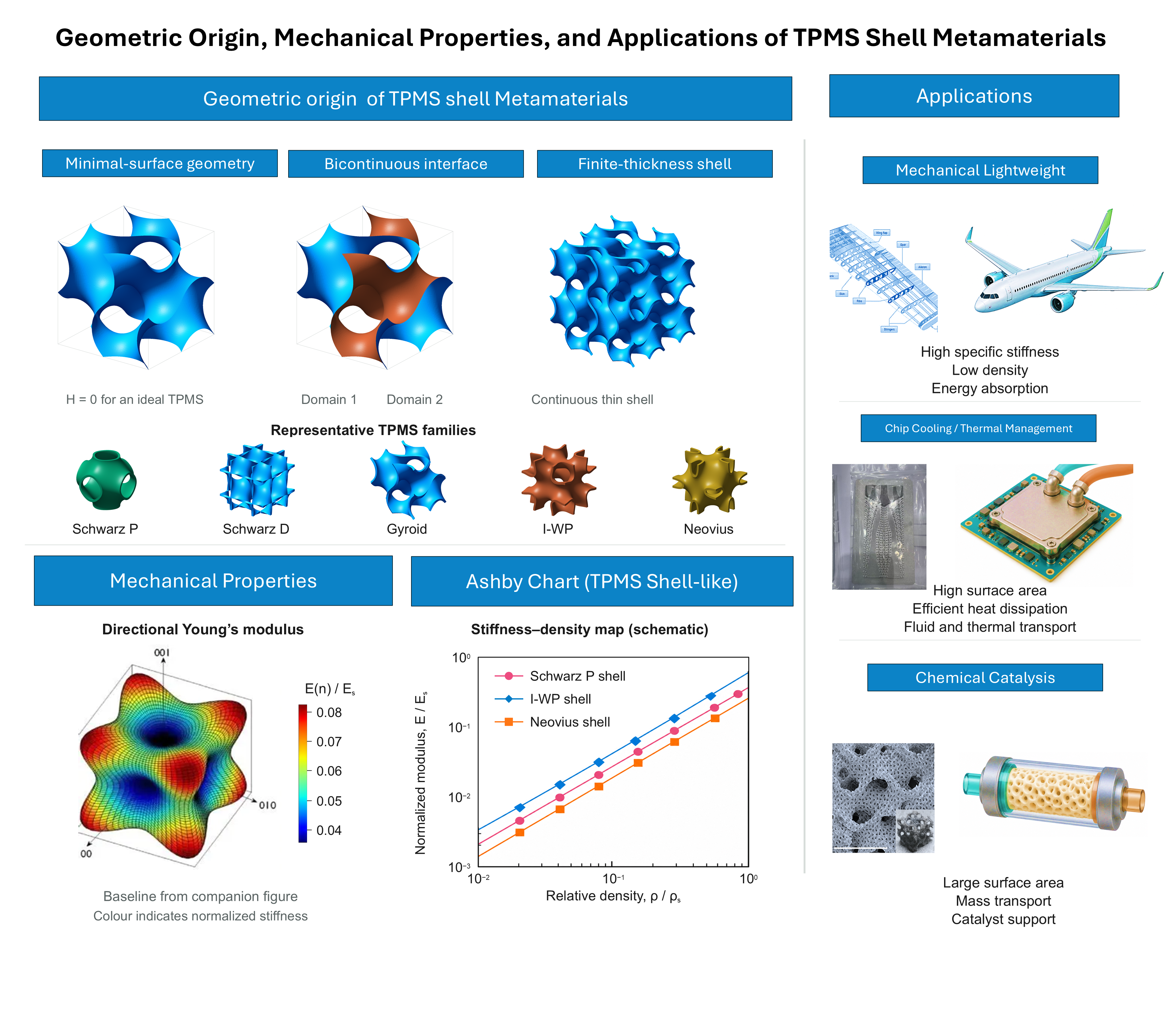}
    \caption{
    Overview of the geometric origin, mechanical characteristics, and representative applications of TPMS shell metamaterials. TPMS shell architectures originate from smooth periodic minimal surfaces and exhibit interconnected geometries that enable favorable mechanical performance and multifunctional applications, including lightweight structures, thermal management, and catalysis\citep{wang2025harnessing,wang2022achieving,fly1232014c919,basavalingappa2017modeling}.
    }
    \label{fig:tpms_intro}
        
\end{figure}

To provide additional context for the structural systems considered in this work, we first briefly summarize the geometric origin, mechanical characteristics, and representative applications of TPMS shell metamaterials. Figure~\ref{fig:tpms_intro} illustrates that TPMS geometries originate from periodic minimal surfaces that partition space into interpenetrating domains. By introducing a finite thickness around the minimal surface, these geometries can be converted into shell-based metamaterials with smooth and highly interconnected architectures. Such structural characteristics provide favorable stiffness-to-weight performance while maintaining large surface-area-to-volume ratios and continuous transport pathways. These properties have motivated applications ranging from lightweight load-bearing structures and energy absorption to thermal management, fluid transport, and catalytic systems.

\subsection{Model Architecture}
\label{sec:model_architecture}

Our model adopts a 3D adaptation of DMAE backbone, which is built upon a Vision Transformer (ViT) encoder–decoder architecture. Given a single-channel TPMS volume \(X\in\mathbb{R}^{1\times64\times64\times64}\), the input is partitioned into non-overlapping \(4\times4\times4\) volumetric patches, resulting in a \(16\times16\times16\) patch grid with \(N=4096\) tokens. Patch embedding is implemented using a 3D convolution with kernel size and stride of 4, and each patch is projected to a \(D=192\)-dimensional embedding. A fixed 3D sinusoidal positional embedding is subsequently added to preserve the spatial location of each volumetric patch.

Following the DMAE/MAE encoding paradigm, masked patches are hiden before the encoder, and only visible patch tokens are processed by the ViT backbone. The encoder consists of six Transformer blocks with an embedding dimension of 192 and six self-attention heads; the feed-forward network in each block expands the feature dimension from 192 to 768 before projecting it back to 192. Each Transformer block uses GELU activation in the feed-forward network, followed by a final LayerNorm after the encoder. For a masking ratio \(r\), the encoder therefore operates on \((1-r)N\) visible tokens and produces:

$$
Z_{\mathrm{vis}}\in\mathbb{R}^{(1-r)N\times192}.
$$

The reconstruction branch follows an asymmetric Transformer decoder. The encoded visible tokens are first projected from 192 to a decoder dimension of 128. Learnable mask tokens are inserted at the removed patch locations, and the complete sequence is restored to its original spatial ordering. After adding 3D sinusoidal positional embeddings, the sequence is processed by 4 Transformer decoder blocks with 4 attention heads and an MLP dimension of 512. A final linear prediction layer maps each decoder token to \(4^3=64\) scalar values, corresponding to one volumetric patch, thereby reconstructing the original \(64\times64\times64\) input volume.

In our topology-aware extension, the DMAE backbone is augmented with an auxiliary persistent-homology prediction head attached to the encoder representation. The shared ViT encoder is jointly optimized by the masked reconstruction and topology-prediction objectives,

$$
\mathcal{L}_{\mathrm{total}}
=
\mathcal{L}_{\mathrm{rec}}
+
\lambda_{\mathrm{PD}}\mathcal{L}_{\mathrm{topo}}.
$$

For downstream representation evaluation, the full volume without masking or noise corruption is processed by the pretrained ViT encoder. The resulting patch-token features are globally averaged to obtain a \(192\)-dimensional structure-level representation, which is subsequently \(L_2\)-normalized before downstream evaluation.

\subsection{Persistent homology for TPMS shell}
\label{sec:PH_TPMS}

We use PH to characterize the multiscale global topology of three-dimensional TPMS structures. Unlike local geometric descriptors, such as surface position, thickness, and curvature, PH captures how topological features—including connected components, tunnels, and  voids—emerge and disappear along a nested sequence of spatial thresholds. PDs summarize these features by encoding their birth, death, and persistence across filtration scales.

\paragraph{From signed-distance fields to cubical complexes.}

Each TPMS structure is represented by an SDF tensor $\tS\in\R^{N\times N\times N}$ sampled on a regular three-dimensional domain, where each entry $\etS_{i,j,k}$ represents the signed distance from the voxel at location $(i,j,k)$ to the TPMS solid--void interface. Negative values correspond to solid regions, whereas positive values correspond to void regions. In our experiments, all SDFs are sampled at a resolution of $N=64$.

To enable persistent-homology analysis, we construct a cubical complex $\sK$ from the regular SDF grid using GUDHI~\citep{gudhi_cubical}. The complex is composed of cubical cells with dimensions $d\in\{0,1,2,3\}$. The three-dimensional cells $c^{(3)}_{i,j,k}$ correspond to cubes, while the lower-dimensional cells $c^{(2)}$, $c^{(1)}$, and $c^{(0)}$ denote faces, edges, and vertices, respectively. Each $c^{(3)}_{i,j,k}\in\sK$ is associated with the SDF entry $\etS_{i,j,k}$.

We define the filtration field by taking the absolute value of the SDF,
\begin{equation}
\etF_{i,j,k}
=
\lvert \etS_{i,j,k} \rvert.
\end{equation}
Here, $\etF_{i,j,k}$ denotes the unsigned distance from the voxel to the solid--void interface. We define the filtration function $f:\sK\rightarrow\R$ to assign a filtration value to each cell in $\sK$, with $f(c^{(3)}_{i,j,k})=\etF_{i,j,k}$ for each cube $c^{(3)}_{i,j,k}$. For a lower-dimensional cell $c^{(d)}$, where $d\in\{0,1,2\}$, its filtration value is set to the minimum value among all incident three-dimensional cubes containing it, i.e., $f(c^{(d)})=\min_{c^{(3)}_{i,j,k}\supseteq c^{(d)}} f(c^{(3)}_{i,j,k})$. Such an assignment ensures closure under inclusion, so whenever a cube enters the filtration, all of its boundary cells are already present.

For a threshold $t\geq0$, the sublevel complex $\sK_t$ contains all cells $c^{(d)}\in\sK$ satisfying $f(c^{(d)})\leq t$, where $d\in\{0,1,2,3\}$. As $t$ increases, cells with larger filtration values are progressively included, corresponding to regions farther from the solid--void interface. The resulting sublevel complexes form a nested filtration,
\begin{equation}
\sK_{t_1}
\subseteq
\sK_{t_2}
\subseteq
\cdots
\subseteq
\sK_{t_m},
\qquad
0\leq t_1<t_2<\cdots<t_m.
\end{equation}
At each filtration level $t$, let $C_q(\sK_t;\mathbb{Z}_2)$ denote the $q$-dimensional chain group over the finite field $\mathbb{Z}_2=\{0,1\}$, where arithmetic is performed modulo 2. The group $C_q(\sK_t;\mathbb{Z}_2)$ consists of linear combinations of the $q$-dimensional cells in $\sK_t$, with $q\in\{0,1,2,3\}$. The boundary operator $\partial_q$ maps each $q$-dimensional chain to its $(q-1)$-dimensional boundary,
\begin{equation}
\partial_q:
C_q(\sK_t;\mathbb{Z}_2)
\rightarrow
C_{q-1}(\sK_t;\mathbb{Z}_2).
\end{equation}
Here, $\partial_q$ encodes the incidence relations between cells of adjacent dimensions; for example, the boundary of a face consists of its edges, whereas the boundary of a cube consists of its faces. A $q$-dimensional cycle is a chain with zero boundary, whereas a $q$-dimensional boundary is a chain that itself forms the boundary of a $(q+1)$-dimensional chain. The corresponding cycle and boundary groups are defined as
\begin{equation}
Z_q(\sK_t)
=
\ker(\partial_q),
\qquad
B_q(\sK_t)
=
\operatorname{im}(\partial_{q+1}),
\end{equation}
where $\ker(\partial_q)$ contains the $q$-dimensional chains mapped to zero by $\partial_q$, and $\operatorname{im}(\partial_{q+1})$ contains the $q$-dimensional chains generated as boundaries of $(q+1)$-dimensional chains.

The $q$-th homology group is then defined as
\begin{equation}
H_q(\sK_t;\mathbb{Z}_2)
=
Z_q(\sK_t) / B_q(\sK_t).
\end{equation}
The quotient removes cycles that are boundaries of higher-dimensional chains, leaving independent homological features represented by $H_q(\sK_t;\mathbb{Z}_2)$~\citep{moor2020topological,bleile2022persistent,edelsbrunner2002topological}.

\paragraph{Topological features.}
At each filtration level $t$, the number of independent homological features in dimension $q$ is quantified by the Betti number
\begin{equation}
\beta_q(t)
=
\dim_{\mathbb{Z}_2} H_q(\sK_t;\mathbb{Z}_2).
\end{equation}
For the three-dimensional cubical complexes considered here, $\beta_0(t)$, $\beta_1(t)$, and $\beta_2(t)$ count connected components, independent tunnels, and voids present at filtration level $t$, respectively~\citep{edelsbrunner2002topological}.

\paragraph{Persistence diagrams.}
Persistent homology tracks the appearance and disappearance of these topological features as the filtration threshold increases. Although a three-dimensional cubical complex can in principle support homology dimensions up to $q=3$, we consider only $q\in\{0,1,2\}$. Here, $H_0$, $H_1$, and $H_2$ characterize connected components, tunnels, and voids, respectively, whereas $H_3$ corresponds to closed three-dimensional cycles without boundary, which do not arise in the finite TPMS domains considered here.

A feature in homology dimension $q\in\{0,1,2\}$ is represented by a birth--death pair $(b_i^{(q)},d_i^{(q)})$, where $b_i^{(q)}$ and $d_i^{(q)}$ denote the filtration thresholds at which the $i$-th feature appears and disappears, respectively. Its persistence is defined as
\begin{equation}
p_i^{(q)}
=
d_i^{(q)}-b_i^{(q)}.
\end{equation}
Only finite persistence intervals with persistence exceeding the prescribed threshold $\epsilon=0.005$ are retained. For each homology dimension, the retained birth--death pairs form the corresponding persistence diagram,
\begin{equation}
\mathrm{PD}_q
=
\{
(b_i^{(q)},d_i^{(q)})
\}_{i=1}^{n_q},
\qquad
q\in\{0,1,2\},
\end{equation}
where $n_q$ denotes the number of retained topological features in homology dimension $q$. Accordingly, $\mathrm{PD}_0$, $\mathrm{PD}_1$, and $\mathrm{PD}_2$ summarize the multiscale persistence of connected components, tunnels, and voids, respectively. Features with larger persistence remain present over a broader range of filtration thresholds and therefore correspond to more persistent topological structures~\citep{zomorodian2005computing,cohensteiner2007stability}.

\subsection{Topological diversity of the dataset}
\label{sec:PH_diversity}
To illustrate the topological diversity of the dataset, we selected 20 representative TPMS samples spanning a broad range of persistence-diagram characteristics. The samples were chosen to exhibit distinct combinations of topological feature counts in $H_0$, $H_1$, and $H_2$, together with substantial variation in feature persistence and birth--death distributions. This selection therefore covers diverse connected-component, tunnel, and cavity structures, providing a representative set of geometries with clearly differentiated persistent-homology signatures.
\begin{figure}[H]
    \centering
    \includegraphics[width=1\textwidth]{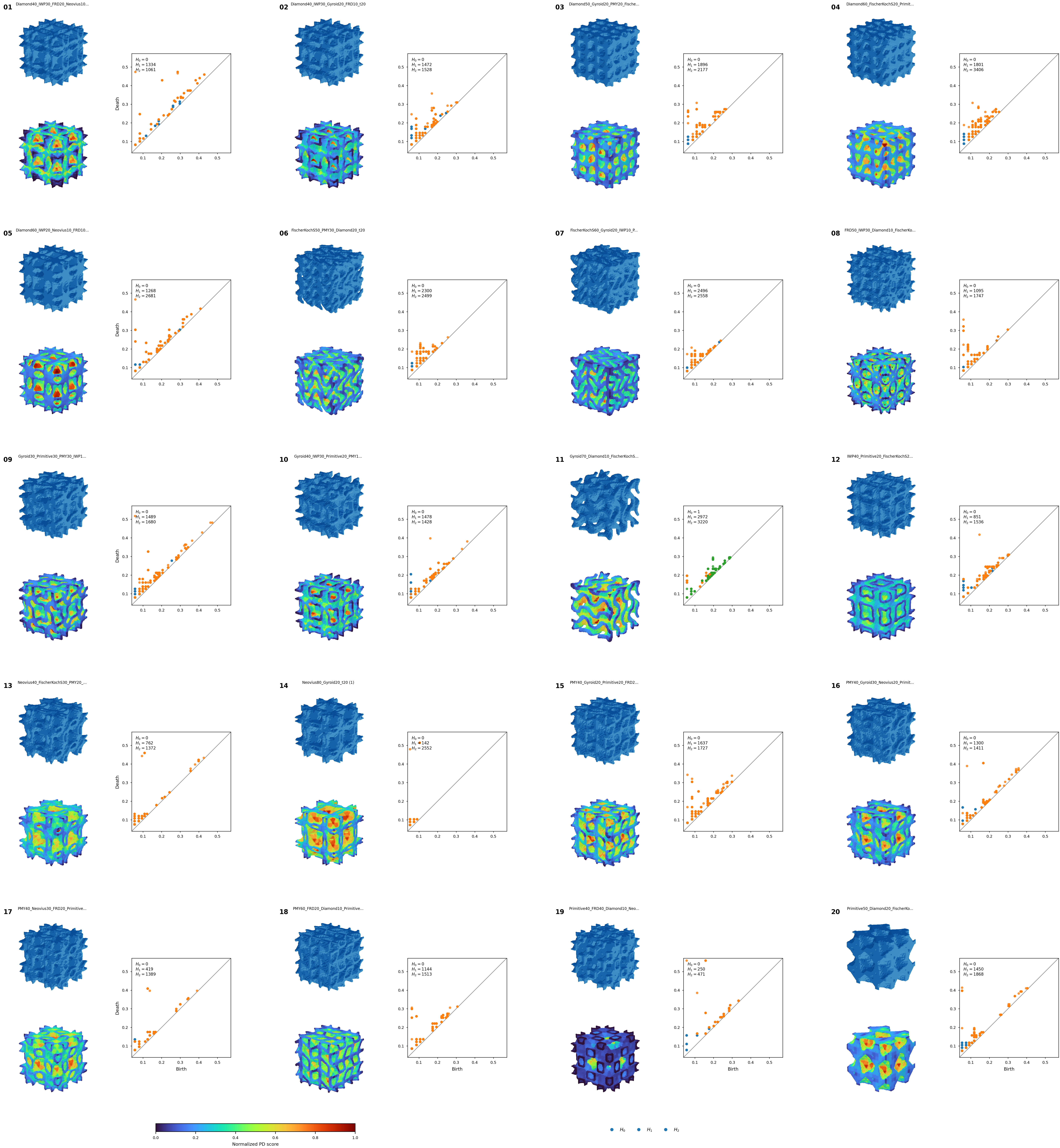}
    \caption{
\textbf{Representative TPMS structures with diverse persistent-homology signatures.}
Twenty TPMS samples were selected to span distinct persistence-diagram characteristics. Each voxelized geometry is shown alongside its corresponding persistence diagram, with the selected set exhibiting different combinations of $H_0$, $H_1$, and $H_2$ feature counts as well as substantial variation in their birth--death distributions. This selection provides a representative visualization of the broad topological diversity present in the TPMS dataset.
}
    \label{fig:PH_diversity}
        
\end{figure}

\subsection{Finite-element homogenization}
\label{sec:fe_homogenization}
The effective elastic properties of each TPMS structure were evaluated using a voxel-based finite-element homogenization method under small-strain linear elasticity. Each geometry was discretized on a regular voxel grid and treated as a periodic unit cell. Periodic boundary conditions were imposed through periodic node mapping, and six independent macroscopic unit-strain states, corresponding to the Voigt components $11$, $22$, $33$, $23$, $31$, and $12$, were applied to determine the homogenized elastic response. The resulting stress responses were assembled into the effective $6\times6$ stiffness matrix $\mathbf{C}$, which contains 21 independent components due to elastic symmetry. The same voxel discretization, constituent-material properties, and homogenization settings were used for the ground-truth, reconstructed, and generated structures, ensuring that differences in the computed mechanical response arose from changes in geometry rather than from the numerical setup. Representative TPMS geometries and their corresponding homogenized stiffness matrices $\mathbf{C}$ are shown below to illustrate the mechanical responses obtained from the finite-element homogenization procedure.
\label{sec:supp_tpms_background}
\begin{figure}[H]
    \centering
    \includegraphics[width=\textwidth]{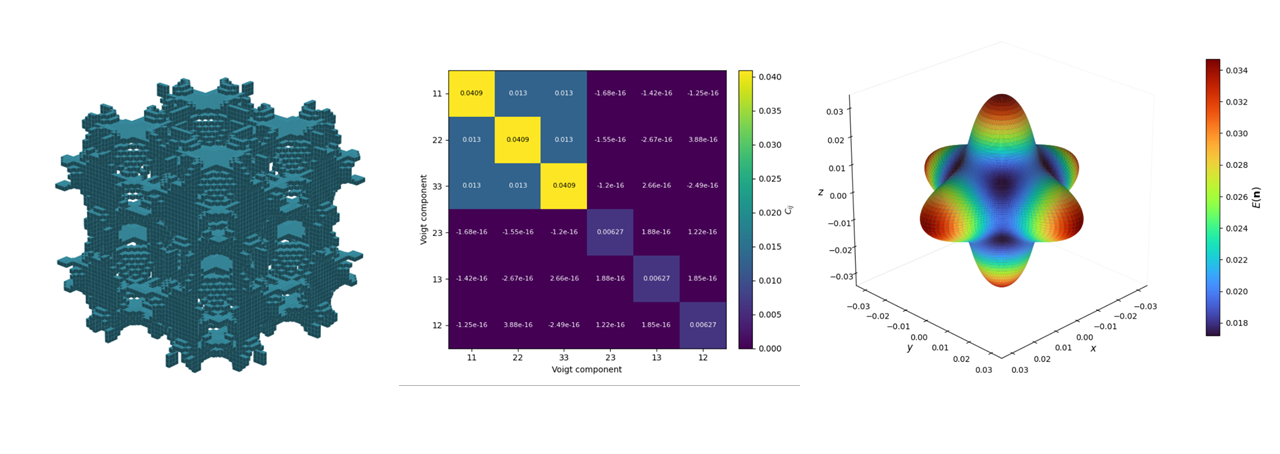}
    \caption{
    \textbf{Representative finite-element homogenization example.}
    A representative TPMS structure is shown together with its homogenized elastic response obtained from the voxel-based finite-element analysis. The corresponding $6\times6$ effective stiffness matrix $\mathbf{C}$ summarizes the anisotropic elastic properties of the structure, while the three-dimensional directional Young's modulus surface $E(\mathbf{n})$, derived from the compliance matrix $\mathbf{S}=\mathbf{C}^{-1}$, visualizes the variation in effective stiffness with loading direction.
    }
    \label{fig:fem_example}
\end{figure}

\subsection{Downstream Tasks}
\label{sec:downstream_appendix}

\subsubsection{Task 2: $C_{11}$ Elastic-Property Classification}

We evaluate whether the frozen representations separate structures according to their mechanical-property levels by formulating $C_{11}$ prediction as a three-class classification problem. The continuous $C_{11}$ values are divided into \textit{Low}, \textit{Medium}, and \textit{High} groups based only on the training-set statistics. Two lightweight probes are applied to the frozen latent representations: a $k$-nearest-neighbor ($k$NN) classifier with $k=15$ and a linear logistic-regression classifier. No encoder parameters are updated during downstream evaluation.

For both probes, we report balanced accuracy and macro-F1. For compactness, balanced accuracy is denoted by $\mathrm{Acc}$ in Table~\ref{tab:c11_classification}. For $C$ classes, balanced accuracy is defined as

\begin{equation}
\mathrm{Acc}
=
\frac{1}{C}
\sum_{c=1}^{C}
\frac{\mathrm{TP}_c}
{\mathrm{TP}_c+\mathrm{FN}_c},
\end{equation}

where $\mathrm{TP}_c$ and $\mathrm{FN}_c$ denote the numbers of true-positive and false-negative samples for class $c$, respectively. Thus, $\mathrm{Acc}$ corresponds to the mean recall across classes and prevents a dominant property group from disproportionately affecting the reported performance.

For each class $c$, precision and recall are defined as

\begin{equation}
P_c
=
\frac{\mathrm{TP}_c}
{\mathrm{TP}_c+\mathrm{FP}_c},
\qquad
R_c
=
\frac{\mathrm{TP}_c}
{\mathrm{TP}_c+\mathrm{FN}_c},
\end{equation}

where $\mathrm{FP}_c$ denotes the number of false-positive samples. The class-wise F1 score is

\begin{equation}
F_{1,c}
=
\frac{2P_cR_c}{P_c+R_c},
\end{equation}

and the macro-F1 score is

\begin{equation}
\mathrm{F}
=
\frac{1}{C}
\sum_{c=1}^{C} F_{1,c}.
\end{equation}

Macro-F1 assigns equal importance to all property groups and therefore jointly reflects class-wise precision and recall. The subscripts $k\mathrm{NN}$ and $\mathrm{Lin}$ distinguish the $k$NN and linear probes, respectively.

As reported in Table~\ref{tab:c11_classification}, our method achieves the best performance across all four classification metrics on the Interpolation TPMS dataset, reaching $\mathrm{F}{k\mathrm{NN}}=0.887$ and $\mathrm{F}{\mathrm{Lin}}=0.925$.
On MetaDB, it achieves the highest linear-probe performance, with $\mathrm{F}_{\mathrm{Lin}}=0.964$, while remaining within $0.003$ of the best $k$NN results.
These results indicate that the learned representations retain mechanically relevant information that can be effectively predicted by simple downstream probes.

\label{sec:supp_downstream}


\begin{table*}[t]
\caption{
\textbf{Task 2 -- Low/medium/high $C_{11}$ elastic-property classification.}
$\mathrm{Acc}$ denotes balanced accuracy and $\mathrm{F}$ denotes macro-F1.
The subscripts $k\mathrm{NN}$ and $\mathrm{Lin}$ correspond to the $k$NN and linear probes, respectively.
Higher is better for all metrics.
}
\label{tab:c11_classification}

\begin{center}
\small
\setlength{\tabcolsep}{4pt}
\renewcommand{\arraystretch}{1.08}

\begin{tabular}{lcccc|cccc}
\hline

\textsc{Model}
& \multicolumn{4}{c|}{\textsc{MetaDB}}
& \multicolumn{4}{c}{\textsc{Interpolation TPMS}}
\\

\cline{2-5}
\cline{6-9}

&
$\mathrm{Acc}_{k\mathrm{NN}}\uparrow$
&
$\mathrm{F}_{k\mathrm{NN}}\uparrow$
&
$\mathrm{Acc}_{\mathrm{Lin}}\uparrow$
&
$\mathrm{F}_{\mathrm{Lin}}\uparrow$
&
$\mathrm{Acc}_{k\mathrm{NN}}\uparrow$
&
$\mathrm{F}_{k\mathrm{NN}}\uparrow$
&
$\mathrm{Acc}_{\mathrm{Lin}}\uparrow$
&
$\mathrm{F}_{\mathrm{Lin}}\uparrow$
\\

\hline

\multicolumn{9}{l}{\textbf{\textit{General representation learning}}}
\\

MVAE
& 0.509 & 0.509 & 0.631 & 0.631
& 0.719 & 0.717 & 0.512 & 0.511
\\

SVAE
& 0.766 & 0.765 & 0.793 & 0.793
& 0.863 & 0.863 & 0.698 & 0.697
\\

VICReg
& 0.853 & 0.853 & 0.913 & 0.913
& 0.799 & 0.801 & 0.835 & 0.837
\\

I-JEPA
& 0.902 & 0.902 & 0.911 & 0.912
& 0.831 & 0.833 & 0.863 & 0.864
\\

\hline

\multicolumn{9}{l}{\textbf{\textit{Masking strategy baselines}}}
\\

MAE
& 0.907 & 0.907 & 0.901 & 0.901
& 0.883 & 0.883 & 0.896 & 0.896
\\

DMAE
& 0.893 & 0.894 & 0.907 & 0.907
& 0.870 & 0.871 & 0.923 & 0.923
\\

ColorMAE
& 0.952 & 0.952 & 0.957 & 0.956
& 0.852 & 0.853 & 0.918 & 0.918
\\

DaM
& 0.948 & 0.948 & 0.952 & 0.952
& 0.857 & 0.858 & 0.912 & 0.912
\\

\hline

\multicolumn{9}{l}{\textbf{\textit{Topology-informed baseline}}}
\\

ScalableTopo
& \textbf{0.954}
& \textbf{0.954}
& 0.960
& 0.957
& 0.856
& 0.856
& 0.916
& 0.916
\\

\hline

\textsc{Ours}
& 0.951
& 0.951
& \textbf{0.964}
& \textbf{0.964}
& \textbf{0.887}
& \textbf{0.887}
& \textbf{0.924}
& \textbf{0.925}
\\

\hline
\end{tabular}
\end{center}
\end{table*}

\subsubsection{Task 3: $C_{11}$ Property Clustering}

We next examine whether mechanically similar structures are naturally organized together in the frozen latent space without using $C_{11}$ labels to train the representation. The resulting latent clusters are evaluated from two complementary perspectives: their ability to explain continuous variations in $C_{11}$ and their agreement with the discrete \textit{Low}/\textit{Medium}/\textit{High} property groups.

We first report the correlation ratio $\eta^2$, which measures the fraction of the total $C_{11}$ variation explained by the latent clustering. Let $y_i$ denote the continuous $C_{11}$ value of sample $i$, $\bar{y}$ denote the global mean, and let cluster $k$ contain $n_k$ samples with mean property value $\bar{y}_k$. The correlation ratio is defined as

\begin{equation}
\eta^2
=
\frac{
\sum_{k=1}^{K}
n_k
\left(
\bar{y}_k-\bar{y}
\right)^2
}{
\sum_{i=1}^{N}
\left(
y_i-\bar{y}
\right)^2
},
\end{equation}

where $N$ is the total number of samples and $K$ is the number of clusters. The numerator measures the between-cluster variation in $C_{11}$, whereas the denominator measures the total variation across the dataset. Therefore, a larger $\eta^2$ indicates that samples assigned to different latent clusters exhibit more distinct mechanical-property distributions.

Because cluster indices are arbitrary, direct comparison between predicted cluster IDs and reference property classes is not meaningful. We therefore determine an optimal one-to-one mapping $\pi$ between cluster indices and reference classes and compute the matched clustering accuracy as

\begin{equation}
\mathrm{Acc}_{\mathrm{Match}}
=
\max_{\pi}
\frac{1}{N}
\sum_{i=1}^{N}
\mathbb{I}
\left[
\pi(z_i)=y_i^{\mathrm{cls}}
\right],
\end{equation}

where $z_i$ is the predicted cluster assignment, $y_i^{\mathrm{cls}}$ is the reference \textit{Low}/\textit{Medium}/\textit{High} property label, $\pi$ denotes the cluster-to-class assignment, and $\mathbb{I}[\cdot]$ is the indicator function. The optimal mapping removes the arbitrary permutation of cluster identities, so $\mathrm{Acc}_{\mathrm{Match}}$ reflects the fraction of samples whose latent clusters agree with the corresponding property groups.

After applying the same optimal cluster-to-class mapping, we additionally compute the macro-F1 score,

\begin{equation}
\mathrm{F}_{\mathrm{Macro}}
=
\frac{1}{C}
\sum_{c=1}^{C}
\frac{2P_cR_c}{P_c+R_c},
\end{equation}

where $P_c$ and $R_c$ denote precision and recall for property class $c$. Unlike matched accuracy, $\mathrm{F}_{\mathrm{Macro}}$ weights all property classes equally and is therefore more sensitive to whether the clustering quality is consistent across different $C_{11}$ ranges.

As shown in Table~\ref{tab:c11_clustering}, our method achieves the best clustering performance across all three metrics on both datasets.
On MetaDB, it reaches $\eta^2=0.323$, $\mathrm{Acc}_{\mathrm{Match}}=0.660$, and $\mathrm{F}_{\mathrm{Macro}}=0.673$, outperforming the strongest competing baseline for each metric.
On the Interpolation TPMS dataset, it similarly achieves the highest $\eta^2=0.412$, $\mathrm{Acc}_{\mathrm{Match}}=0.582$, and $\mathrm{F}_{\mathrm{Macro}}=0.576$, indicating that the learned latent space is more strongly organized according to continuous $C_{11}$ variation and the corresponding property groups.

\begin{table*}[t]
\caption{
\textbf{Task 3 -- $C_{11}$ property clustering.}
$\eta^2$ measures the fraction of continuous $C_{11}$ variation explained by the latent clusters.
$\mathrm{Acc}_{\mathrm{Match}}$ and $\mathrm{F}_{\mathrm{Macro}}$ evaluate agreement with the discrete property groups after optimal cluster-to-class matching.
Higher is better for all metrics.
}
\label{tab:c11_clustering}

\begin{center}
\small
\setlength{\tabcolsep}{5pt}
\renewcommand{\arraystretch}{1.08}

\begin{tabular}{lccc|ccc}
\hline

\textsc{Model}
& \multicolumn{3}{c|}{\textsc{MetaDB}}
& \multicolumn{3}{c}{\textsc{Interpolation TPMS}}
\\

\cline{2-4}
\cline{5-7}

&
$\eta^2\uparrow$
&
$\mathrm{Acc}_{\mathrm{Match}}\uparrow$
&
$\mathrm{F}_{\mathrm{Macro}}\uparrow$
&
$\eta^2\uparrow$
&
$\mathrm{Acc}_{\mathrm{Match}}\uparrow$
&
$\mathrm{F}_{\mathrm{Macro}}\uparrow$
\\

\hline

\multicolumn{7}{l}{\textbf{\textit{General representation learning}}}
\\

MVAE
& 0.036 & 0.044 & 0.041
& 0.014 & 0.372 & 0.367
\\

SVAE
& 0.127 & 0.519 & 0.436
& 0.049 & 0.399 & 0.394
\\

VICReg
& 0.068 & 0.457 & 0.457
& 0.020 & 0.382 & 0.377
\\

I-JEPA
& 0.014 & 0.393 & 0.291
& 0.081 & 0.370 & 0.366
\\

\hline

\multicolumn{7}{l}{\textbf{\textit{Masking strategy baselines}}}
\\

MAE
& 0.104
& 0.433
& 0.401
& 0.258
& 0.391
& 0.380
\\

DMAE
& 0.147
& 0.504
& 0.436
& 0.272
& 0.410
& 0.383
\\

ColorMAE
& 0.255
& 0.630
& 0.603
& 0.162
& 0.413
& 0.393
\\

DaM
& 0.196
& 0.514
& 0.497
& 0.177
& 0.446
& 0.414
\\

\hline

\multicolumn{7}{l}{\textbf{\textit{Topology-informed baseline}}}
\\

ScalableTopo
& 0.199
& 0.543
& 0.533
& 0.158
& 0.422
& 0.395
\\

\hline

\textsc{Ours}
& \textbf{0.323}
& \textbf{0.660}
& \textbf{0.673}
& \textbf{0.412}
& \textbf{0.582}
& \textbf{0.576}
\\

\hline
\end{tabular}
\end{center}
\end{table*}

\subsubsection{Task 4: $C_{11}$ Property Regression}

Finally, we evaluate whether the continuous mechanical-property value can be quantitatively recovered from the frozen representation. A regression probe is trained on the latent features to predict $C_{11}$, while the representation encoder remains fixed. We evaluate the predictions using mean absolute error (MAE), root mean squared error (RMSE), and the coefficient of determination ($R^2$).

Let $y_i$ and $\hat{y}_i$ denote the ground-truth and predicted $C_{11}$ values of sample $i$, respectively. The mean absolute error is defined as

\begin{equation}
\mathrm{MAE}
=
\frac{1}{N}
\sum_{i=1}^{N}
\left|
y_i-\hat{y}_i
\right|,
\end{equation}

where $N$ is the number of evaluated samples. MAE measures the average absolute prediction error in the same units as $C_{11}$ and weights all errors linearly.

The root mean squared error is defined as

\begin{equation}
\mathrm{RMSE}
=
\sqrt{
\frac{1}{N}
\sum_{i=1}^{N}
\left(
y_i-\hat{y}_i
\right)^2
}.
\end{equation}

Because the prediction errors are squared before averaging, RMSE places a larger penalty on samples with large regression errors and is therefore more sensitive to outliers than MAE.

We further report the coefficient of determination,

\begin{equation}
R^2
=
1-
\frac{
\sum_{i=1}^{N}
\left(
y_i-\hat{y}_i
\right)^2
}{
\sum_{i=1}^{N}
\left(
y_i-\bar{y}
\right)^2
},
\end{equation}

where $\bar{y}$ denotes the mean ground-truth $C_{11}$ value. The numerator represents the residual sum of squares of the regression predictions, while the denominator represents the total variation in the ground-truth properties. An $R^2$ value closer to 1 indicates that a larger fraction of the variation in $C_{11}$ can be recovered from the frozen latent representation.

As shown in Table~\ref{tab:c11_regression}, our method achieves competitive regression performance on MetaDB, with $\mathrm{MAE}=0.009$, $\mathrm{RMSE}=0.047$, and $R^2=0.899$.
On the Interpolation TPMS dataset, it achieves the lowest MAE ($0.003$) and RMSE ($0.005$), while matching the highest $R^2$ of $0.993$.
These results show that the learned representation preserves fine-grained continuous mechanical-property information in addition to the discrete and clustering-level organization evaluated in Tasks~2 and~3.

\begin{table*}[t]
\caption{
\textbf{Task 4 -- $C_{11}$ property regression.}
MAE measures the average absolute prediction error, RMSE places greater emphasis on large errors, and $R^2$ measures the fraction of $C_{11}$ variation explained by the regression prediction.
Lower is better for MAE and RMSE, while higher is better for $R^2$.
}
\label{tab:c11_regression}

\begin{center}
\small
\setlength{\tabcolsep}{5pt}
\renewcommand{\arraystretch}{1.08}

\begin{tabular}{lccc|ccc}
\hline

\textsc{Model}
& \multicolumn{3}{c|}{\textsc{MetaDB}}
& \multicolumn{3}{c}{\textsc{Interpolation TPMS}}
\\

\cline{2-4}
\cline{5-7}

&
$\mathrm{MAE}\downarrow$
&
$\mathrm{RMSE}\downarrow$
&
$R^2\uparrow$
&
$\mathrm{MAE}\downarrow$
&
$\mathrm{RMSE}\downarrow$
&
$R^2\uparrow$
\\

\hline

\multicolumn{7}{l}{\textbf{\textit{General representation learning}}}
\\

MVAE
& 0.049 & 0.078 & 0.246
& 0.039 & 0.050 & 0.110
\\

SVAE
& 0.028 & 0.051 & 0.471
& 0.024 & 0.031 & 0.616
\\

VICReg
& 0.011 & 0.049 & 0.903
& 0.010 & 0.014 & 0.935
\\

I-JEPA
& 0.010 & 0.048 & 0.907
& 0.007 & 0.010 & 0.965
\\

\hline

\multicolumn{7}{l}{\textbf{\textit{Masking strategy baselines}}}
\\

MAE
& 0.010
& 0.047
& 0.909
& 0.005
& 0.007
& 0.984
\\

DMAE
& 0.009
& 0.050
& 0.899
& \textbf{0.003}
& \textbf{0.005}
& \textbf{0.993}
\\

ColorMAE
& \textbf{0.007}
& 0.046
& 0.915
& \textbf{0.003}
& \textbf{0.005}
& 0.992
\\

DaM
& 0.010
& 0.057
& 0.869
& 0.004
& \textbf{0.005}
& 0.991
\\

\hline

\multicolumn{7}{l}{\textbf{\textit{Topology-informed baseline}}}
\\

ScalableTopo
& 0.008
& \textbf{0.041}
& \textbf{0.932}
& \textbf{0.003}
& \textbf{0.005}
& 0.992
\\

\hline

\textsc{Ours}
& 0.009
& 0.047
& 0.899
& \textbf{0.003}
& \textbf{0.005}
& \textbf{0.993}
\\

\hline
\end{tabular}
\end{center}
\end{table*}

\begin{table*}[t]
\centering
\caption{
\textbf{Ablation study on Interpolation TPMS and MetaDB.}
Results are reported as mean $\pm$ standard deviation over five random seeds.
For Interpolation TPMS, Task 1 reports strict and relaxed $k$NN family-classification accuracy;
for MetaDB, Task 1 reports macro-F1 from the linear and $k$NN probes.
Task 2 reports $F_{k\mathrm{NN}}$ for $C_{11}$ property classification;
Task 3 reports $\eta^2$, ARI, and NMI for $C_{11}$ property clustering;
and Task 4 reports $R^2$ for $C_{11}$ and the full stiffness tensor $C_{21}$.
Higher is better for all metrics.
}
\label{tab:ablation_full}

\resizebox{\textwidth}{!}{
\begin{tabular}{lcccccccc}
\toprule

\multicolumn{9}{l}{\textit{Interpolation TPMS}} \\

& \multicolumn{2}{c}{Task 1}
& \multicolumn{1}{c}{Task 2}
& \multicolumn{3}{c}{Task 3}
& \multicolumn{2}{c}{Task 4} \\

\cmidrule(lr){2-3}
\cmidrule(lr){4-4}
\cmidrule(lr){5-7}
\cmidrule(lr){8-9}

Ablation
& $\mathrm{Acc}^{\mathrm{Str}}_{k\mathrm{NN}}\uparrow$
& $\mathrm{Acc}^{\mathrm{Rel}}_{k\mathrm{NN}}\uparrow$
& $F_{k\mathrm{NN}}\uparrow$
& $\eta^2\uparrow$
& $\mathrm{ARI}\uparrow$
& $\mathrm{NMI}\uparrow$
& $R^2_{C_{11}}\uparrow$
& $R^2_{C_{21}}\uparrow$ \\

\midrule

w/o mask \& head
& $0.8191 \pm 0.0126$
& $0.9324 \pm 0.0061$
& $0.8710 \pm 0.0068$
& $0.2721 \pm 0.0060$
& $0.138 \pm 0.019$
& $0.124 \pm 0.020$
& $\mathbf{0.9925 \pm 0.0003}$
& $0.9901 \pm 0.0002$ \\

w/o mask
& $0.8218 \pm 0.0077$
& $0.9358 \pm 0.0069$
& $0.8845 \pm 0.0062$
& $0.4045 \pm 0.0317$
& $0.196 \pm 0.041$
& $0.191 \pm 0.046$
& $0.9925 \pm 0.0005$
& $\mathbf{0.9964 \pm 0.0003}$ \\

w/o head
& $0.8232 \pm 0.0155$
& $0.9365 \pm 0.0115$
& $0.8682 \pm 0.0062$
& $0.2508 \pm 0.0328$
& $0.116 \pm 0.040$
& $0.116 \pm 0.030$
& $0.9917 \pm 0.0002$
& $0.9961 \pm 0.0003$ \\

Full
& $\mathbf{0.8309 \pm 0.0074}$
& $\mathbf{0.9391 \pm 0.0073}$
& $\mathbf{0.8870 \pm 0.0049}$
& $\mathbf{0.4116 \pm 0.0363}$
& $\mathbf{0.203 \pm 0.015}$
& $\mathbf{0.213 \pm 0.013}$
& $0.9923 \pm 0.0006$
& $0.9962 \pm 0.0003$ \\

\midrule

\multicolumn{9}{l}{\textit{MetaDB}} \\

& \multicolumn{2}{c}{Task 1}
& \multicolumn{1}{c}{Task 2}
& \multicolumn{3}{c}{Task 3}
& \multicolumn{2}{c}{Task 4} \\

\cmidrule(lr){2-3}
\cmidrule(lr){4-4}
\cmidrule(lr){5-7}
\cmidrule(lr){8-9}

Ablation
& $F_{\mathrm{Lin}}\uparrow$
& $F_{k\mathrm{NN}}\uparrow$
& $F_{k\mathrm{NN}}\uparrow$
& $\eta^2\uparrow$
& $\mathrm{ARI}\uparrow$
& $\mathrm{NMI}\uparrow$
& $R^2_{C_{11}}\uparrow$
& $R^2_{C_{21}}\uparrow$ \\

\midrule

w/o mask \& head
& $0.9070 \pm 0.0016$
& $0.8940 \pm 0.0015$
& $0.8940 \pm 0.0086$
& $0.1470 \pm 0.0050$
& $0.130 \pm 0.029$
& $0.166 \pm 0.034$
& $0.8990 \pm 0.0130$
& $0.8640 \pm 0.0150$ \\

w/o mask
& $0.9872 \pm 0.0019$
& $0.9720 \pm 0.0021$
& $0.9314 \pm 0.0065$
& $0.3146 \pm 0.0088$
& $0.202 \pm 0.052$
& $0.324 \pm 0.047$
& $0.9066 \pm 0.0181$
& $0.9061 \pm 0.0149$ \\

w/o head
& $0.9860 \pm 0.0040$
& $\mathbf{0.9930 \pm 0.0010}$
& $0.9460 \pm 0.0070$
& $0.3150 \pm 0.0110$
& $0.210 \pm 0.057$
& $0.305 \pm 0.068$
& $\mathbf{0.9370 \pm 0.0200}$
& $0.9350 \pm 0.0100$ \\

Full
& $\mathbf{0.9944 \pm 0.0014}$
& $0.9845 \pm 0.0014$
& $\mathbf{0.9513 \pm 0.0040}$
& $\mathbf{0.3228 \pm 0.0130}$
& $\mathbf{0.211 \pm 0.046}$
& $\mathbf{0.331 \pm 0.047}$
& $0.8991 \pm 0.0191$
& $\mathbf{0.9360 \pm 0.0178}$ \\

\bottomrule
\end{tabular}
}
\end{table*}

\subsection{Clustering Metrics in Ablation Study}
\label{sec:clustering_metrics}

In addition to $\eta^2$, we evaluate the alignment between the unsupervised $K$-means clusters and the discretized $C_{11}$ property groups using the Adjusted Rand Index (ARI) and Normalized Mutual Information (NMI).

Let $n_{ij}$ denote the number of samples shared by predicted cluster $i$ and property class $j$, with row sums $a_i=\sum_j n_{ij}$, column sums $b_j=\sum_i n_{ij}$, and total sample size $n$. The ARI is defined as

$$
\mathrm{ARI}
=
\frac{
\displaystyle
\sum_{ij}\binom{n_{ij}}{2}
-
\frac{
\displaystyle
\sum_i\binom{a_i}{2}
\sum_j\binom{b_j}{2}
}{
\binom{n}{2}
}
}{
\displaystyle
\frac{1}{2}
\left[
\sum_i\binom{a_i}{2}
+
\sum_j\binom{b_j}{2}
\right]
-
\frac{
\displaystyle
\sum_i\binom{a_i}{2}
\sum_j\binom{b_j}{2}
}{
\binom{n}{2}
}
}.
$$

ARI corrects the agreement between the two partitions for chance, with larger values indicating stronger correspondence.

For predicted clusters $U$ and property classes $V$, NMI is computed from their mutual information,

$$
I(U;V)
=
\sum_{i,j}
p_{ij}
\log
\frac{p_{ij}}{p_i p_j},
$$

and normalized by the entropies of the two partitions:

$$
\mathrm{NMI}
=
\frac{I(U;V)}
{\sqrt{H(U)H(V)}},
$$

where

$$
H(U)=-\sum_i p_i\log p_i,
\qquad
H(V)=-\sum_j p_j\log p_j.
$$

NMI ranges from 0 to 1, with higher values indicating stronger agreement between the clustering structure and the $C_{11}$ property groups.

\subsection{Latent-Space Alignment Metrics}
\label{sec:latent_alignment_metrics}

We evaluate whether pairwise distances in the learned latent space reflect topology- and mechanics-related differences between TPMS structures. For each model, the latent vectors are standardized using statistics computed from the corresponding training split. For a held-out pair of structures $(i,j)$, the latent distance is defined as
\[
d_z(i,j)
=
\left\|
\widetilde{\mathbf z}_i
-
\widetilde{\mathbf z}_j
\right\|_2,
\]
where $\widetilde{\mathbf z}$ denotes the standardized latent representation.

For each descriptor, we compute a corresponding pairwise distance $d_t(i,j)$ and quantify its alignment with the latent distance using the Spearman rank correlation
\[
\rho_s
=
1-
\frac{6\sum_{m=1}^{M}\left(R_m^{z}-R_m^{t}\right)^2}
{M(M^2-1)},
\]
where $M$ is the number of evaluated structure pairs, and $R_m^{z}$ and $R_m^{t}$ denote the ranks of the latent distance and descriptor distance for pair $m$, respectively. Higher $\rho_s$ indicates stronger correspondence between the learned latent geometry and the corresponding structural descriptor.

\paragraph{PD-Score Alignment.}
Let $s_{i,k}$ denote the PD score of patch $k$ in structure $i$. Following the masking procedure, the scores are normalized within each structure as
\[
\pi_{i,k}
=
\frac{s_{i,k}}
{\sum_{r=1}^{n_p}s_{i,r}},
\]
where $n_p$ is the number of patches. The pairwise PD-score distance is
\[
d_{\mathrm{PD}}(i,j)
=
\left\|
\boldsymbol{\pi}_i
-
\boldsymbol{\pi}_j
\right\|_2.
\]
PD-Score Alignment is then defined as
\[
\rho_{\mathrm{PD}}
=
\rho_s
\left(
d_z,
d_{\mathrm{PD}}
\right).
\]

\paragraph{$H_1$- and $H_2$-Count Alignment.}
Let $n_i^{(q)}$ denote the number of persistence features in homology dimension $q\in\{1,2\}$ for structure $i$. The pairwise count distance is
\[
d_{H_q}(i,j)
=
\left|
n_i^{(q)}
-
n_j^{(q)}
\right|.
\]
The corresponding alignment is
\[
\rho_{H_q}
=
\rho_s
\left(
d_z,
d_{H_q}
\right),
\qquad q\in\{1,2\}.
\]

\paragraph{$C_{11}$-Property Alignment.}
The scalar stiffness component $C_{11}$ is standardized using statistics from the training split, yielding $\widetilde C_{11}$. The pairwise property distance is
\[
d_{C_{11}}(i,j)
=
\left|
\widetilde C_{11,i}
-
\widetilde C_{11,j}
\right|.
\]
The property alignment is defined as
\[
\rho_{C_{11}}
=
\rho_s
\left(
d_z,
d_{C_{11}}
\right).
\]
\subsection{Parameter Sensitivity and Hyperparameter Selection}
\label{sec:param_sensitivity}

We further examine the sensitivity of the proposed framework to three key hyperparameters: the patch size $p$, the masking ratio $\rho$, and the effective-loss weight $\lambda_{\mathrm{eff}}$, which are most relavant to our proposed mechanisms. 
The results on the Interpolation TPMS dataset are summarized in Table~\ref{tab:param_sensitivity}. 
Each parameter is varied independently while the remaining relevant settings are fixed.

For the patch size, $p=4$ achieves the best results in Task~1,2,and 4. The only exception is Task~3, where $p=8$ yields slightly better clustering performance, improving $\eta^2$, $\mathrm{Acc}_{\mathrm{Match}}$, and $F_{\mathrm{Macro}}$ by only $0.0206$, $0.0150$, and $0.0155$, respectively, compared with $p=4$. 
Given these relatively small gains in Task~3 and the consistently better performance of $p=4$ in Tasks~1, 2, and 4, we adopt $p=4$ for the main experiments.

The masking ratio shows a clearer trade-off between family-level structural discrimination and property-related organization of the latent space. 
Increasing the masking ratio from $\rho=0.75$ to $0.85$ improves the four Task~1 metrics by up to $0.0144$, indicating stronger family discrimination under more aggressive masking. 
However, this comes at a substantially larger cost in Task~3: $\eta^2$ decreases from $0.3166$ to $0.1500$, while $\mathrm{Acc}_{\mathrm{Match}}$ and $F_{\mathrm{Macro}}$ decrease from $0.5508$ and $0.5465$ to $0.4409$ and $0.4412$, respectively. 
The $k$NN property-classification F1 in Task~2 also decreases from $0.8793$ to $0.8728$. 
Conversely, lower masking ratios improve some property-related metrics but lead to noticeably weaker Task~1 performance; for example, reducing $\rho$ from $0.75$ to $0.55$ lowers the two linear-probe family metrics by approximately $0.039$. 
We therefore choose $\rho=0.75$ as a balanced operating point, retaining strong structural discrimination without the substantial degradation in clustering observed at $\rho=0.85$.

A similar trade-off is observed for the PD-loss weight $\lambda_{\mathrm{PD}}$. 
While $\lambda_{\mathrm{PD}}=0.01$ gives the best linear family-classification performance, $\lambda_{\mathrm{PD}}=0.05$ improves $\eta^2$, $\mathrm{Acc}_{\mathrm{Match}}$, and $F_{\mathrm{Macro}}$ in Task~3 by $0.3640$, $0.1609$, and $0.1568$, respectively, at the cost of only $0.0212$ and $0.0187$ in the two linear Task~1 metrics. 
Further increasing $\lambda_{\mathrm{PD}}$ to $0.10$ yields only marginal gains in Task~2 and Task~4 while slightly reducing Task~1 and Task~3 performance. 
We therefore select $\lambda_{\mathrm{PD}}=0.05$ as a balanced setting across tasks.

Based on above sensitivity results and analysis, we use $p=4$, $\rho=0.75$, and $\lambda_{\mathrm{PD}}=0.05$ in the main experiments.

\begin{table*}[t]
\centering
\caption{
\textbf{Parameter sensitivity analysis on the Interpolation TPMS dataset.}
We evaluate the sensitivity to masking ratio $\rho$, PD-loss weight
$\lambda_{\mathrm{PD}}$, and patch size $p$.
For the masking-ratio sweep, $\lambda_{\mathrm{PD}}=0.02$ and $p=4$ are fixed;
for the PD-loss-weight sweep, $\rho=0.75$ and $p=4$ are fixed;
and for the patch-size sweep, $\rho=0.75$ and $\lambda_{\mathrm{PD}}=0.05$ are fixed.
Bold values indicate the best performance within each sensitivity block.
$\dagger$ denotes the value adopted in the main configuration.
}
\label{tab:param_sensitivity}

\resizebox{\textwidth}{!}{
\begin{tabular}{llccccccccccc}
\toprule

& &
\multicolumn{4}{c}{Task 1: Family classification} &
\multicolumn{2}{c}{Task 2: $C_{11}$ classification} &
\multicolumn{3}{c}{Task 3: $C_{11}$ clustering} &
\multicolumn{2}{c}{Task 4: Regression}
\\

\cmidrule(lr){3-6}
\cmidrule(lr){7-8}
\cmidrule(lr){9-11}
\cmidrule(lr){12-13}

Parameter
& Value
& $\mathrm{Acc}^{\mathrm{Str}}_{\mathrm{Lin}}\uparrow$
& $\mathrm{Acc}^{\mathrm{Rel}}_{\mathrm{Lin}}\uparrow$
& $\mathrm{Acc}^{\mathrm{Str}}_{k\mathrm{NN}}\uparrow$
& $\mathrm{Acc}^{\mathrm{Rel}}_{k\mathrm{NN}}\uparrow$
& $F_{k\mathrm{NN}}\uparrow$
& $F_{\mathrm{Lin}}\uparrow$
& $\eta^2\uparrow$
& $\mathrm{Acc}_{\mathrm{Match}}\uparrow$
& $F_{\mathrm{Macro}}\uparrow$
& $R^2_{C_{11}}\uparrow$
& $R^2_{C_{21}}\uparrow$
\\

\midrule


\multirow{4}{*}{Mask ratio $\rho$}
& 0.55
& 0.8376
& 0.8484
& 0.8275
& 0.9317
& \textbf{0.8869}
& \textbf{0.9218}
& \textbf{0.3901}
& 0.5459
& 0.5349
& \textbf{0.99225}
& 0.99628
\\

& 0.65
& 0.8606
& 0.8750
& 0.8272
& 0.9360
& 0.8846
& 0.9157
& 0.3134
& 0.5493
& \textbf{0.5473}
& 0.99185
& \textbf{0.99636}
\\

& $0.75^\dagger$
& 0.8764
& 0.8879
& 0.8347
& 0.9396
& 0.8793
& 0.9075
& 0.3166
& \textbf{0.5508}
& 0.5465
& 0.99190
& 0.99633
\\

& 0.85
& \textbf{0.8908}
& \textbf{0.9023}
& \textbf{0.8491}
& \textbf{0.9450}
& 0.8728
& 0.9108
& 0.1500
& 0.4409
& 0.4412
& 0.99204
& 0.99613
\\

\midrule

\multirow{5}{*}{$\lambda_{\mathrm{PD}}$}

& 0.01
& \textbf{0.8872}
& \textbf{0.8994}
& 0.8279
& 0.9310
& 0.8778
& \textbf{0.9210}
& 0.0972
& 0.4518
& 0.4529
& 0.99184
& 0.99610
\\

& 0.02
& 0.8764
& 0.8879
& \textbf{0.8347}
& \textbf{0.9396}
& 0.8793
& 0.9075
& 0.3166
& 0.5508
& 0.5465
& 0.99190
& \textbf{0.99633}
\\

& $0.05^\dagger$
& 0.8660
& 0.8807
& 0.8315
& 0.9364
& 0.8847
& 0.9165
& \textbf{0.4612}
& \textbf{0.6127}
& \textbf{0.6097}
& 0.99194
& 0.99620
\\

& 0.10
& 0.8591
& 0.8768
& 0.8225
& 0.9299
& \textbf{0.8891}
& 0.9162
& 0.4560
& 0.6075
& 0.6055
& \textbf{0.99221}
& 0.99604
\\

& 0.20
& 0.8365
& 0.8545
& 0.8178
& 0.9271
& 0.8887
& 0.9116
& 0.3129
& 0.5445
& 0.5381
& 0.99206
& 0.99583
\\

\midrule

\multirow{3}{*}{Patch size $p$}

& $4^\dagger$
& \textbf{0.8660}
& \textbf{0.8807}
& \textbf{0.8315}
& \textbf{0.9364}
& \textbf{0.8847}
& \textbf{0.9165}
& 0.4612
& 0.6127
& 0.6097
& \textbf{0.99194}
& \textbf{0.99620}
\\

& 8
& 0.8383
& 0.8548
& 0.8246
& 0.9332
& 0.8789
& 0.9137
& \textbf{0.4818}
& \textbf{0.6277}
& \textbf{0.6252}
& 0.99110
& 0.99587
\\

& 16
& 0.8146
& 0.8311
& 0.8124
& 0.9238
& 0.8810
& 0.9131
& 0.4550
& 0.6137
& 0.6087
& 0.99010
& 0.99567
\\

\bottomrule
\end{tabular}
}
\end{table*}

\begin{table*}[t]
\centering
\caption{
\textbf{Parameter sensitivity analysis on the MetaDB dataset.}
We evaluate the sensitivity to the PD-loss weight $\lambda_{\mathrm{PD}}$,
while fixing the masking ratio to $\rho=0.75$ and patch size to $p=4$.
Bold values indicate the best performance within the sensitivity sweep.
$\dagger$ denotes the value adopted in the main configuration.
}
\label{tab:param_sensitivity_metadb}

\resizebox{\textwidth}{!}{
\begin{tabular}{llccccccccccc}
\toprule

& &
\multicolumn{2}{c}{Task 1: Family classification} &
\multicolumn{4}{c}{Task 2: $C_{11}$ classification} &
\multicolumn{3}{c}{Task 3: $C_{11}$ clustering} &
\multicolumn{2}{c}{Task 4: Regression}
\\

\cmidrule(lr){3-4}
\cmidrule(lr){5-8}
\cmidrule(lr){9-11}
\cmidrule(lr){12-13}

Parameter
& Value
& $F_{\mathrm{Lin}}\uparrow$
& $F_{k\mathrm{NN}}\uparrow$
& $\mathrm{BA}_{\mathrm{Lin}}\uparrow$
& $F_{\mathrm{Lin}}\uparrow$
& $\mathrm{BA}_{k\mathrm{NN}}\uparrow$
& $F_{k\mathrm{NN}}\uparrow$
& $\eta^2\uparrow$
& $\mathrm{Acc}_{\mathrm{Match}}\uparrow$
& $F_{\mathrm{Macro}}\uparrow$
& $R^2_{C_{11}}\uparrow$
& $R^2_{C_{21}}\uparrow$
\\

\midrule

\multirow{14}{*}{$\lambda_{\mathrm{PD}}$}

& 0.005
& \textbf{0.9945}
& 0.9844
& 0.9569
& 0.9569
& 0.9513
& 0.9512
& \textbf{0.3235}
& 0.6593
& 0.6723
& 0.9026
& 0.9169
\\

& $0.010^\dagger$
& 0.9944
& 0.9845
& \textbf{0.9641}
& \textbf{0.9640}
& 0.9513
& 0.9512
& 0.3228
& 0.6598
& 0.6728
& 0.8991
& \textbf{0.9360}
\\

& 0.015
& 0.9930
& \textbf{0.9877}
& 0.9600
& 0.9599
& 0.9539
& 0.9537
& 0.3217
& 0.6598
& 0.6728
& 0.8565
& 0.9033
\\

& 0.020
& 0.9931
& 0.9840
& 0.9635
& 0.9635
& 0.9539
& 0.9537
& 0.3202
& 0.6687
& 0.6809
& 0.9007
& 0.9192
\\

& 0.030
& 0.9937
& 0.9827
& 0.9560
& 0.9560
& 0.9503
& 0.9502
& 0.3073
& 0.5811
& 0.5603
& 0.8753
& 0.9122
\\

& 0.100
& 0.9924
& 0.9859
& 0.9559
& 0.9559
& 0.9513
& 0.9511
& 0.3073
& 0.5801
& 0.5595
& 0.8826
& 0.8981
\\

& 0.150
& 0.9911
& 0.9842
& 0.9600
& 0.9600
& 0.9523
& 0.9522
& 0.3012
& 0.5274
& 0.4989
& \textbf{0.9031}
& 0.9004
\\

& 0.200
& 0.9927
& 0.9848
& 0.9575
& 0.9574
& 0.9503
& 0.9501
& 0.3204
& \textbf{0.6703}
& \textbf{0.6823}
& 0.8898
& 0.9125
\\

& 0.225
& 0.9924
& 0.9869
& 0.9569
& 0.9569
& \textbf{0.9544}
& \textbf{0.9542}
& 0.3057
& 0.5157
& 0.4999
& 0.8849
& 0.9106
\\

& 0.250
& 0.9921
& 0.9868
& 0.9590
& 0.9589
& 0.9518
& 0.9517
& 0.2510
& 0.4522
& 0.4389
& 0.8858
& 0.8995
\\

& 0.275
& 0.9890
& 0.9854
& 0.9560
& 0.9559
& 0.9533
& 0.9532
& 0.2002
& 0.5125
& 0.4799
& 0.8947
& 0.9034
\\

& 0.300
& 0.9900
& 0.9863
& 0.9566
& 0.9564
& 0.9528
& 0.9527
& 0.2032
& 0.5217
& 0.5044
& 0.9009
& 0.9078
\\

& 0.350
& 0.9906
& 0.9848
& 0.9605
& 0.9605
& 0.9508
& 0.9506
& 0.1078
& 0.4711
& 0.4550
& 0.8520
& 0.9039
\\

& 0.400
& 0.9923
& 0.9838
& 0.9584
& 0.9585
& 0.9498
& 0.9497
& 0.1092
& 0.4636
& 0.4492
& 0.9024
& 0.8950
\\

\bottomrule
\end{tabular}
}
\end{table*}

\end{document}

%% file: math_commands.tex
\usepackage{amsmath,amsfonts,bm}

\def\eqref#1{equation~\ref{#1}}

\def\1{\bm{1}}

\DeclareMathAlphabet{\mathsfit}{\encodingdefault}{\sfdefault}{m}{sl}
\SetMathAlphabet{\mathsfit}{bold}{\encodingdefault}{\sfdefault}{bx}{n}
\newcommand{\tens}[1]{\bm{\mathsfit{#1}}}

\def\tP{{\tens{P}}}

\def\tS{{\tens{S}}}

\def\sK{{\mathbb{K}}}

\def\sM{{\mathbb{M}}}

\def\sV{{\mathbb{V}}}

\newcommand{\etens}[1]{\mathsfit{#1}}

\def\etF{{\etens{F}}}

\def\etS{{\etens{S}}}

\newcommand{\R}{\mathbb{R}}

